\documentclass[journal]{IEEEtran}
\pdfoutput=1
\usepackage{amsmath}
\usepackage{amsfonts}
\usepackage{amssymb}

\usepackage{amsthm}
\usepackage{mathtools}
\usepackage{tabularx,booktabs}
\usepackage{graphicx}
\usepackage{subfigure}
\usepackage{enumerate}
\usepackage{float}
\usepackage{url}
\usepackage{verbatim}
\usepackage{cite}
\usepackage{diagbox}
\usepackage{multirow}
\usepackage{makecell}
\usepackage{algorithm}  
\usepackage{algorithmicx}  
\usepackage{algpseudocode} 
\usepackage{bbding}
\usepackage{utfsym}
\usepackage{bm}
\usepackage{cases}
\usepackage[linkcolor=black,citecolor=black,urlcolor=black,colorlinks=true]{hyperref}
\usepackage{tikz}
\usepackage{etoolbox}
\usepackage{graphicx}
\usepackage{latexsym}
\usepackage{pifont}
\usepackage{threeparttable}

\newcommand{\circled}[2][]{\tikz[baseline=(char.base)]
	{\node[shape = circle, draw, inner sep = 1pt]
		(char) {\phantom{\ifblank{#1}{#2}{#1}}};%
		\node at (char.center) {\makebox[0pt][c]{#2}};}}
\robustify{\circled}

\graphicspath{{figures/}}
\IEEEoverridecommandlockouts

\allowdisplaybreaks[4]

\newcommand{\df}[1]{\mathrm{d}{#1}}

\newcommand{\norm}[1]{\Vert{#1}\Vert}

\author{Yongchao Wang$^{*1}$, Junjie Wang$^{*2}$, Xiaobin Zhou$^{2}$, Tiankai Yang$^{2}$, Chao Xu$^{2}$, and Fei Gao$^{\dagger2}$
    \thanks{ $^*$Indicates equal contribution. }
	\thanks{$^\dagger$Corresponding author: {\tt\small fgaoaa@zju.edu.cn}.}
    \thanks{
    $^1$School of Aeronautic Science and Engineering, Beihang University, Beijing 100191, China(e-mail: wangyongchao@buaa.edu.cn).
	}
	\thanks{
	$^2$Key Laboratory of Industrial Control Technology, Institute of Cyber-Systems and Control, Zhejiang University, Hangzhou 310027, China, and the Huzhou Institute, Zhejiang University, Huzhou 313000, China. (e-mail: wwwangjunjie@zju.edu.cn)
	}
    \thanks{
	This work was supported by the National Natural Science Foundation of China under grant no. 62403419, and the Fundamental Research Funds for the Central Universities.
	}
}

\title{\LARGE \bf Safe and Agile Transportation of Cable-Suspended Payload via \\ Multiple Aerial Robots}

\begin{document}
    \maketitle

\begin{abstract}
Transporting a heavy payload using multiple aerial robots (MARs) is an efficient manner to extend the load capacity of a single aerial robot. However, existing schemes for the multiple aerial robots transportation system (MARTS) still lack the capability to generate a collision-free and dynamically feasible trajectory in real-time and further track an agile trajectory especially when there are no sensors available to measure the states of payload and cable. Therefore, they are limited to low-agility transportation in simple environments. To bridge the gap, we propose complete planning and control schemes for the MARTS, achieving safe and agile aerial transportation (SAAT) of a cable-suspended payload in complex environments. Flatness maps for the aerial robot considering the complete kinematical constraint and the dynamical coupling between each aerial robot and payload are derived. To improve the responsiveness for the generation of the safe, dynamically feasible, and agile trajectory in complex environments, a real-time spatio-temporal trajectory planning scheme is proposed for the MARTS. Besides, we break away from the reliance on the state measurement for both the payload and cable, as well as the closed-loop control for the payload, and propose a fully distributed control scheme to track the agile trajectory that is robust against imprecise payload mass and non-point mass payload. The proposed schemes are extensively validated through benchmark comparisons, ablation studies, and simulations. Finally, extensive real-world experiments are conducted on a MARTS integrated by three aerial robots with onboard computers and sensors. The result validates the efficiency and robustness of our proposed schemes for SAAT in complex environments.  

% Indirect position representation is adopted to eliminate the kinematical constraint between each aerial robot and payload introduced by the cable,

% In addition, existing control schemes that either directly introduce state measurement for the payload or ignore dynamical coupling between the payload and the AR are fundamentally constraining on trajectory tracking performance.

\end{abstract}


\section{Introduction}
\label{sec:intro}
\IEEEPARstart{A}{erial} transportation is more time-efficient compared to ground transportation, as it is unaffected by terrain and traffic congestion. In areas such as logistics, medical rescue, and dangerous fire and disaster scenes, various time-sensitive supplies and equipment can be delivered by aerial transportation. Transporting a payload via just one aerial robot has been widely studied \cite{sreenath2013geometric, foehn2017fast,tang2018aggressive, son2020real, zeng2020differential,belkhale2021model, yu2022aggressive, li2023autotrans,  wang2024impact}. However, the limited load capacity of a single aerial robot greatly restricts its application in different fields.

\begin{figure}[t]
    \begin{center}
        \includegraphics[width=1.0\columnwidth]{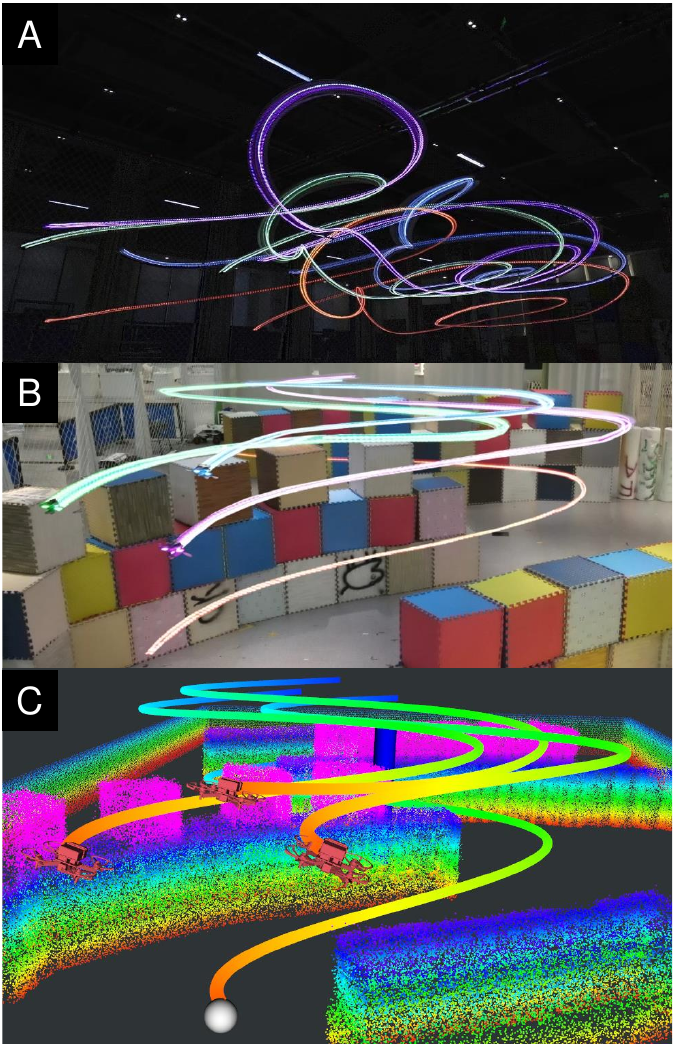}
    \end{center}
    \vspace{-0.3 cm}
    \caption{Simulations and real-world experiments of our MARTS. (A) Agile transportation in free space, approaching the limit of thrust can be provided by the practical aerial robot in the MARTS. (B) and (C) display the actual snapshot and rviz simulation of the SAAT in a complex environment. Please watch our attached videos for more information at: 
    https://youtu.be/deD2wD673iI.}
    \vspace{-0.5 cm}
    \label{fig:head}
\end{figure}

% https://youtu.be/LuQ2O9mXElE

% One approach to transport a payload that exceeds the upper limit of an AR's load capability is to design a new and larger AR. However, this is not an economical option since in addition to expensive research and development costs, the manufacturing costs of a AR increase faster than the efficient load capability as the size increases. The second approach is to utilize  

Using MARs to transport a payload collaboratively can enhance the load capability of an aerial robot with arbitrary size. The payload can be attached to the MARs via rigid connections such as grippers \cite{mellinger2013cooperative}, electromagnets \cite{loianno2017cooperative}, and rigid rods \cite{wang2018cooperative}, making up a large underactuated system whose acceleration can only be provided by first regulating the attitude of the whole system, thereby the agility reduces since the payload increases the system's rotational inertia. Therefore, using cables to suspend the payload has attracted extensive interest from the robotics community \cite{bernard2011autonomous, masone2016cooperative, prajapati2020human, li2024rotortm} and is preferred as it not only eliminates the need for additional actuators and reduces structural complexity, thus increasing the load capacity of the system, but also allows agile transportation while minimizing the payload's attitude. In this paper, we consider the cable-suspended payload transportation problem via the MARTS.

% Complex and diverse mission scenes as well as limited endurance necessitate high autonomy and mission efficiency, which demands the MARs to agilely transport the payload and avoid collisions with all the obstacles.

Speedy transportation of a payload in complex environments relies on a planner to generate a safe and agile trajectory, as well as a control scheme to track this trajectory. However, these are not easy works since the dynamical coupling and the kinematical constraint between each aerial robot and payload introduced by the tight cable complicate the problem. Several practical challenges are analyzed as follows.   

The first challenge is how to ensure the \textit{dynamical feasibility} of the agile trajectory planned for the MARTS with the dynamical coupling and the kinematical constraint. Due to the limited battery life, increasing the agility of trajectory is necessary to improve the mission efficiency \cite{kaufmann2023champion}. However, as the trajectory becomes more agile, the state and control input of the aerial robot may transiently or persistently exceed the upper limits of its maneuvering capability, causing large tracking errors and thus resulting in system divergence and even crashes. Moreover, the dynamical coupling can exacerbate this trend. Therefore, flatness maps considering the dynamical coupling and the kinematical constraint existing in MARTS need to be derived to calculate the state and control input of each aerial robot, on which delicate constraints can be enforced to ensure the dynamical feasibility. 

The second challenge is how to track an agile trajectory without state measurements of payload and cable even when there are uncertainties on the payload. Generally, a closed-loop control for the payload needs to be achieved in MARTS's control scheme. However, the controllers containing this closed-loop control are not suitable for the MARTS to track an agile trajectory. One reason is that the closed-loop control for the payload essentially acts as a centralized part of the MARTS's control scheme and the control law would be allocated and transmitted to each aerial robot, which not only induces disadvantageous time-delay but also inevitably propagates the oscillation in the control law to each aerial robot's desired position. Besides, from the mission perspective, it is not convenient to obtain precise priori information about the payload's mass. Moreover, it is not realistic to assemble sensors on the cables and especially the payload to measure their states, and the realistic payload cannot be strictly mass points. Therefore, a robust control scheme to uncertainties on the payload, getting rid of the reliance on the closed-loop control for payload as well as state measurements for both payload and cable, is necessary. 

% without measuring the payload's state even when there are uncertainties on the payload. From the mission perspective, it is not convenient to obtain precise priori information about the mass of payload. Besides, it is not realistic to assemble sensors on the payload to measure its state and attach the cables to the payload's center of mass (CoM) exactly. Moreover, these are no direct forces to control the payload, and the cable' force is indirectly provided by first regulating AR' attitude and net thrust with inevitable error and lag, which fundamentally increases the difficulty of control. Therefore, a high-accuracy and robust control architecture for agile trajectory tracking is crucial for AT. 

Apart from the above requirements, the safety of the trajectory and responsiveness of the planner must also be considered for the deployment of MARTS. Safety is crucial, as collisions involving the entire system with obstacles or reciprocal collisions between aerial robots can lead to crashes. Responsiveness is also important, as rapid trajectory generation enhances deployment speed and improves the system's ability to react to sudden changes, such as target alterations.

Based on the above analysis, we propose complete planning and control schemes for the MARTS. To the extent of our knowledge, this is the first work that achieves the real-time trajectory generation for SAAT of the cable-suspended payload by the MARTS in complex environments and achieves the transportation of a payload approaching the maximum agility of the MARTS by pushing the aerial robot's thrust to its limit. In this work, we indirectly represent the aerial robot's position via the cable's directional vector and the payload's position, such that the kinematical constraint can be eliminated. Firstly, we derive the flatness maps considering the dynamical coupling and the kinematical constraint existing in MARTS to obtain the high-order states of the aerial robot. Secondly, we propose an efficient lightweight unconstrained trajectory planning scheme to fastly generate a trajectory based on a trajectory class represented by sparse spatio-temporal parameters. This planning scheme constructs not only the obstacle avoidance constraint for each component of MARTS and the reciprocal avoidance constraint for each pair of aerial robots to ensure safety, but also the delicate dynamical feasibility constraints to ensure the agile trajectory is dynamically executable. Besides, it also constructs vectorial constraints for each cable's force which are dexterously eliminated by diffeomorphisms to avoid the undesirable local-optimum trajectory. Thirdly, we propose a fully distributed control scheme to track the agile trajectory relying solely on the state measurements of aerial robots and is robust against the imprecise estimation of payload's mass and non-point mass payloads. Finally, we deploy our planning and control schemes on a practical MARTS consisting of three aerial robots and design various experiments to verify our schemes. Partial results are shown in Fig. 1. Contributions of this paper are listed as:
\begin{enumerate}
    \item Flatness maps that consider the dynamical coupling and the kinematical constraint between each aerial robot and payload are derived based on the indirect position representation for the aerial robot.

    \item A real-time trajectory planner for the MARTS, which considers the dynamical feasibility of aerial robots, the safety of the entire system in complex environments, and the finite vectorial ranges of cables, to generate safe and agile trajectories.
    
    \item A robust and distributed control scheme to track the agile trajectory, which doesn't rely on the closed-loop control for payload and state measurements for both payload and cable and is robust against model uncertainties on the payload.
    
    \item A variety of simulations and real-world experiments that validate the effectiveness of proposed planning and control schemes on a practical MARTS. Moreover, we open-source the code to facilitate further development of MARTS by the community.  
\end{enumerate}

% Efficient trajectory optimization enables high-frequency replanning even in embedded processors with severely limited resources.

% \begin{enumerate}
%     \item A series of differentiable planning metrics are constructed on the flat mappings derived by indirect position representation for the drone, which enable safety and dynamical feasibility in aggressive flight efficiently.

%     \item A real-time trajectory optimization framework considering full state dynamics and collision constraints for safe and agile transportation in complex environments.
    
%     \item A robust, distributed control framework relying just on the sensors on the drones for tracking a maneuvering trajectory with high control accuracy even when there are model uncertainties on the payload.
    
%     \item A variety of simulations and real-world tests that validate the proposed methods on a platform with three quadrotors. Moreover, we open-source the code to facilitate further development of MARS by the community.  
% \end{enumerate}

%carry out intensive collision detections along all the cables' swept surfaces, the ARs' trajectories as well as the payload's trajectory, and further

% keep the system away from obstacles
% with frequent map queries is inevitable at the expense of an increased computational burden. 

\section{Related Works}
\label{sec: related works}
\subsection{Motion Planning for MARTS}
Several works \cite{michael2011cooperative, fink2011planning, jiang2012inverse} have investigated collaborative manipulation and transportation of a payload by MARTS based on the quasi-static assumption. However, these methods are severely limited because the inertial forces in agile transportation cannot be ignored simply. 

Manubens et al. \cite{manubens2013motion} use the Transition-based RRT to search for a collision-free non-smooth path for the MARTS. However, only simple dynamics are taken into account, so it is equally unsuitable for agile transportation. Sreenath et al. \cite{sreenath2013dynamics} reveal that the MARTS is differentially flat when all the cables are tight and plan a smooth trajectory via minimizing the $6^{th}$ derivative of flat output. However, the planning scheme neglects the safety of MARTS in complex environments and the dynamical feasibility of the planned trajectory. Besides, the deformation efficiency of the minimum-snap trajectory class \cite{mellinger2011minimum}, on which this scheme relies cannot guarantee the real-time performance of trajectory planning. Jackson et al. \cite{jackson2020scalable} consider the safety of MARTS in simple environments by simplifying the obstacle as a cylinder and the dynamical feasibility. They parallelize the solving of the optimization problem to mitigate the explosion of optimization time as the number of aerial robots increases. Nevertheless, it still takes a few seconds before the convergence even when the MARTS just consists of three aerial robots. Besides, the quality of the optimized trajectory is suspectable since they just enforce an incomplete kinematical constraint (relative distance constraint) between each aerial robot and payload and neglect the high-order constraints on relative velocity and acceleration. Sun et al. \cite{sun2023nonlinear} consider the complete kinematical constraint and dynamical feasibility of trajectory and propose a real-time nonlinear model predictive control (NMPC) with high scalability to the number of aerial robots. However, they neglect the safety of MARTS in complex environments. Wahba et al.\cite{wahba2023kinodynamic} consider not only the complete kinematical constraint but also the dynamical feasibility and the safety of MARTS in complex environments. However, they still do not address the real-time problem of trajectory planning.

\subsection{Control for MARTS}
Lee et al.\cite{lee2013geometric, lee2017geometric} propose geometric controllers for MARTS under an assumption that both the payload's state and the cable's state are known, considering the point-mass payload and the rigid body payload respectively. Li et al. \cite{li2021cooperative} use monocular vision and inertial module to estimate the cable's state and partial payload's states in a distributed manner, but the payload's acceleration and jerk required to calculate the geometric control law are still neglected. All these controllers \cite{lee2013geometric, lee2017geometric, li2021cooperative} cannot guarantee the avoidance of reciprocal collision between the aerial robots since they use a determinate minimum-norm principle to allocate each cable's desired force vector for payload's control. To ensure reciprocal avoidance, optimization-based \cite{geng2020cooperative} and NMPC-based \cite{li2023nonlinear} force allocation methods are proposed via exploiting the redundancy in the null space to ensure the minimum safe distance between the aerial robots. However, all the controllers \cite{lee2013geometric, lee2017geometric, li2021cooperative, geng2020cooperative, li2023nonlinear} rely on special sensors or marks mounted on the payload to estimate or measure the payload's state, which is not convenient in practical deployment. Besides, the payload's acceleration and jerk, as well as the cable's angular rate introduce noise to the control law of the payload and lead to numerical instability. Moreover, all these controllers can be classified into centralized controllers for all the force allocations are centralized. The time-delay introduced by centralized computation and data transmission will limit the performances of these controllers, especially for agile transportation. Wahba et al. \cite{wahba2024efficient} propose a distributed force allocation method, in which each aerial robot can update its desired force vector with a low frequency via optimizing the same problem existing only one global minima with the same input such that consistent results can be ensured. However, this method also does not fundamentally address the dependence on the payload's state and the detrimental effect of noise on control.

A common denominator of all the controllers \cite{lee2013geometric, lee2017geometric, li2021cooperative, geng2020cooperative, li2023nonlinear, wahba2024efficient} mentioned above is that they pursue the closed-loop control for payload. However, the oscillation existing in the payload's tracking error will directly lead to the oscillation of the aerial robot's desired position. Moreover, the control force for the payload can only be provided indirectly once the aerial robot has adjusted its attitude and the thrust to regulate the cable's force vector. Therefore, the response lag and the inevitable control error will fundamentally restrict the performance of these controllers. All these are the reasons why agile transportation is rarely seen in existing works.

\begin{figure*}[ht]
    \begin{center}
        \includegraphics[width=1.88\columnwidth]{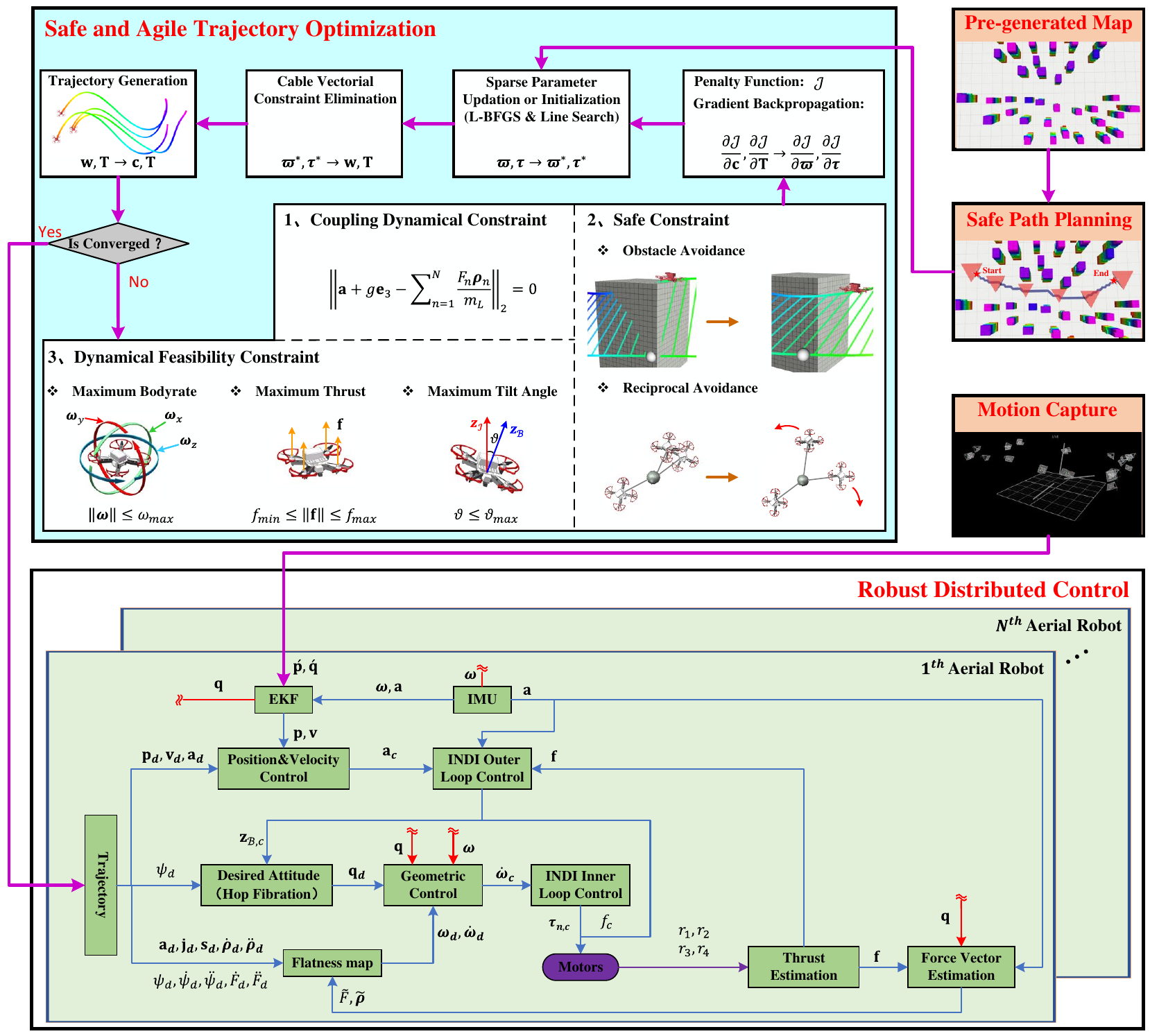}
    \end{center}
    \vspace{-0.3 cm}
    \caption{An overview of our safe and agile trajectory planning scheme and robust distributed control scheme for MARTS.}
    \label{fig:system_architecture}
\end{figure*}

To cope with the drawbacks of these controllers mentioned above, several fully distributed control schemes are proposed \cite{klausen2018cooperative, de2019flexible}. They form a formation to transport the payload and simply treat the cable's force as an external disturbance that can be compensated via an observer. However, these control schemes neglect the negative influence induced by dynamical couplings. Besides, distributed lead-follower control schemes for the beam-like payload are studied in \cite{gassner2017dynamic, tognon2018aerial, tagliabue2019robust}. However, the follower's controller can only react to the leader's motion passively, which leads to a nonsmooth trajectory and is too short-sighted to avoid the reciprocal collision.      

% \begin{figure*}[t]
%     \begin{center}
%         \includegraphics[width=2.0\columnwidth]{figures/Fig2.pdf}
%     \end{center}
%     \vspace{-0.3 cm}
%     \caption{An overview of our complete aerial system with dynamic tracking and perching scheme. The trajectory generation module takes all the requirements mentioned in Sec. \ref{sec:intro} into account and provides spatiotemporal optimal feasible trajectory for stable tracking and dynamic perching.}
%     \vspace{-0.2 cm}
%     \label{fig:system}
% \end{figure*}

\section{System Overview and Preliminaries}
\label{sec:system}
\subsection{System Architecture}
The overall architecture of our planning and control schemes is illustrated in Fig.~\ref{fig:system_architecture}. Fusing LiDAR feature points with IMU data using FAST-lio2 \cite{xu2022fast} package, we construct the environmental map. A system-level global path planning method (Sec.~\ref{sec:safe path finding}) that treats the MARTS as a convex and scalable orthopyramid is designed to find a global safe path based on this map, which acts as the initial value of the sparse parameters for the back-end safe and agile trajectory optimization. 

For each iteration of the safe and agile trajectory optimization, the auxiliary sparse parameters $\bm \varpi^*, \bm \tau^*$ are firstly mapped to actual sparse parameters $\mathbf w, \mathbf T$, which is the unique representation for the trajectory, by diffeomorphisms (Sec.~\ref{subsubsec:Topological Constraint Elimination}) such that the cable's vectorial constraints imposed on the actual sparse parameters can be eliminated directly. Then, a new trajectory can be generated from the actual sparse parameters by calculating a system of linear equations with linear spatio-temporal computational complexity (Sec.~\ref{subsec:traj rep}). Next, we calculate the high-order states and control input using the flatness maps (Sec.~\ref{subsec:Flatness Maps}) such that the quality of trajectory can be evaluated in terms of the safety (Sec.~\ref{subsec:Safety Constraints}), the dynamical feasibility (Sec.~\ref{subsec:Dynamical Feasibility Constraints}) and the coupled dynamical constraint (Sec.~\ref{subsec:Coupling Dynamic Constraint}) specifically existing in our scheme. Besides, the total penalty cost and the gradients w.r.t the trajectory's parameters $\mathbf c, \mathbf T$ are computed consequently. Finally, these gradients are backpropagated in turn to the gradients w.r.t the actual and auxiliary sparse parameters consequently, based on which the auxiliary sparse parameters update to new values (Sec.~\ref{subsec:Unconstrained NLP formation}). The optimized trajectory will be transmitted to each aerial robot when the optimization satisfies the condition of convergence.          
     
We get the odometry for each aerial robot using an extended Kalman filter (EKF) to fuse the position and velocity from the motion capture system with the onboard IMU data. Our robust and distributed control scheme mainly includes a two-loop control method shared by all the aerial robots. The motor's revolutions per minute (RPM) are measured to estimate the thrust and control torque of the aerial robot. The outer loop (Sec.~\ref{subsec:Outer Loop Controller}) adopts the incremental nonlinear dynamic inversion (INDI) \cite{9121690} to get the thrust vector command based on the closed-loop trajectory tracking control law, the filter acceleration, and the estimated thrust vector. The inner loop (Sec.~\ref{subsec:Inner Loop Controller}) adopts the hop fibration rotation factorization to reconstruct the desired attitude and calculate the desired body rate as well as angular acceleration by flatness maps, which also uses INDI to get the control torque command based on the geometric attitude control law, the estimated angular acceleration, and the estimated control torque. Finally, the control commands are transformed into the signals to control each motor's RPM.

\begin{table}[t]
	\renewcommand\arraystretch{1.2}
	\centering
	\caption{Symbol Definition}
	\label{tab:symbol definition}
        \resizebox{\linewidth}{!}{
\begin{tabular}{lp{5cm}}
  \toprule
  Symbol & Definition \\  
  \midrule
  $\mathcal{I}, \mathcal{B}_n$   & world frame, the $n^{th}$ robot's body frame\\
  $\mathbf z_{\mathcal I},\mathbf z_{\mathcal B}^{n} \in \mathbb R^3$  & z-axises of $\mathcal{I}$ and $\mathcal{B}_n$ \\
  $\mathbf e_3$   & unit vector $[0,0,1]^T$ \\
  $m_L \in \mathbb R_{>0}$    &  mass of payload \\
  $m_n \in \mathbb R_{>0}$    &  mass of the $n^{th}$ aerial robot \\
  $\mathbf  J_n \in \mathbb R^{3 \times 3}$    & rotational inertia of the $n^{th}$ aerial robot \\
  $\mathbf p, \mathbf v\in \mathbb R^3$  & position and velocity of payload \\
  $\mathbf p_n, \mathbf v_n\in \mathbb R^3$  & position and velocity of the $n^{th}$ aerial robot\\
  $\mathbf p_n^k \in \mathbb R^3$  & position of the $k^{th}$ sampled point on the $n^{th}$ cable \\
  $\mathbf R_n \in  \mathrm{SO}(3)$  & rotation matrix of the $n^{th}$ aerial robot \\
  $\psi_n \in  \mathbb R$  & yaw angle of the $n^{th}$ aerial robot \\
  $\vartheta_n \in  \mathbb R$  & tile angle of the $n^{th}$ aerial robot, i.e., the angle between $\bm z_{\mathcal B}^{n} $ and $\bm z_{\mathcal I} $ \\
  $\mathbf q_n $  & quaternion of the $n^{th}$ aerial robot \\
  $\bm \omega_n \in \mathbb R^3$  & body rate w.r.t $\mathcal{B}_n$ of the $n^{th}$ aerial robot\\
  $\bm \rho_n \in S^2$  & unit vector of the $n^{th}$ cable w.r.t $\mathcal{I}$, pointing from payload to the $n^{th}$ aerial robot  \\
  $\bm \tau_n \in \mathbb R^{3}$  & control moment of the $n^{th}$ aerial robot w.r.t $\mathcal{B}_n$ \\
  $\mathbf f_n \in \mathbb R^{3}$  &  mass-normalized thrust vector of the $n^{th}$ aerial robot \\
  $f_n \in \mathbb R_{>0}$ & mass-normalized thrust of the $n^{th}$ aerial robot  \\

  $F_n \in \mathbb R_{>0}$  & tension of the $n^{th}$ cable \\
  $\theta_n, \phi_n \in  \mathbb R$  & pitch angle, azimuth angle of the $n^{th}$ cable w.r.t $\mathcal{I}$ \\
  $l \in \mathbb R_{>0}$  & length of the $n^{th}$ cable \\
  $g \in \mathbb R_{>0}$  & gravitational acceleration \\
%   $\tilde{\mathbf x} $  &  estimation of the variable $\mathbf x$\\
  \bottomrule
\end{tabular}}
\end{table}

\subsection{Dynamics Model}
\label{subsec:dynamic model}
\begin{figure}[t]
    \begin{center}
         \includegraphics[width=0.97\columnwidth]{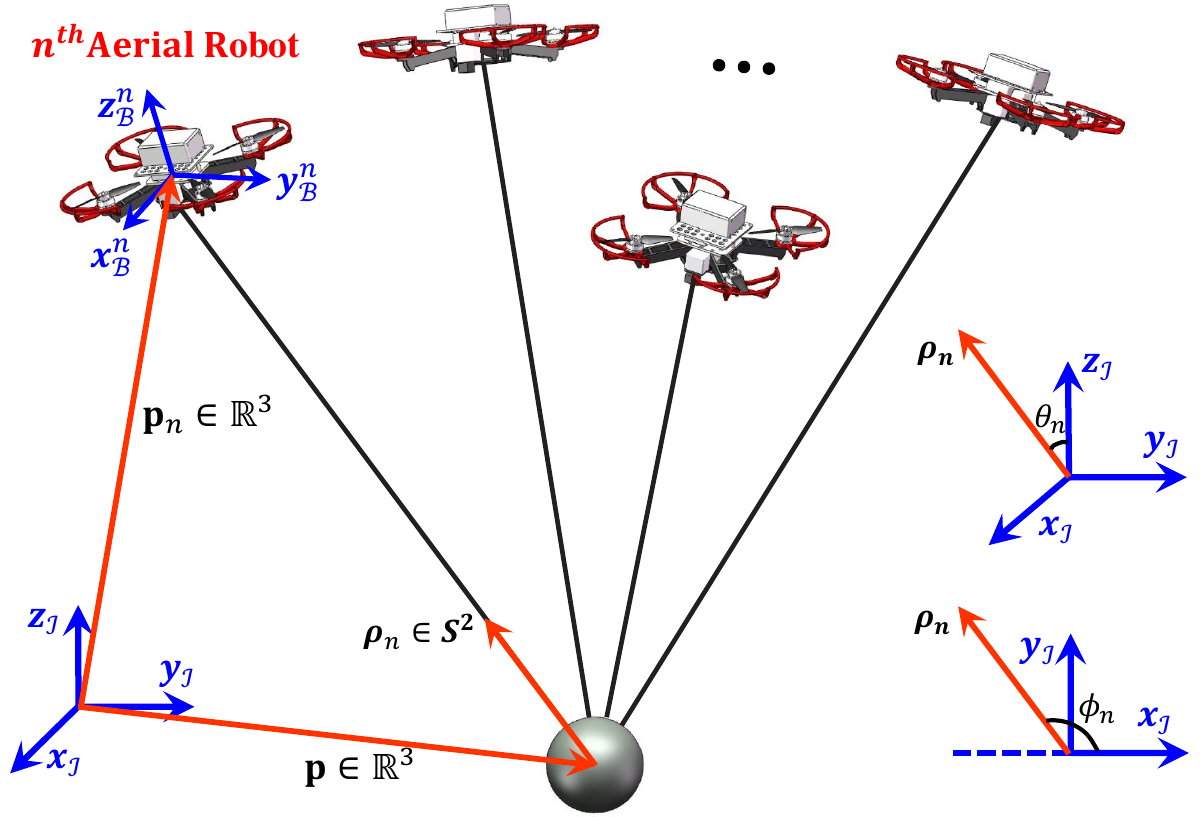}
    \end{center}
    \caption{Illustration of the cable-suspended payload transportation by the MARTS and the definition of $\theta_n, \phi_n$ used to represent $\bm \rho_n$.} \label{fig:system_Illustration}
\end{figure}

In this paper, we consider a point-mass payload transported by a MARTS consisting of $N$ aerial robots, as is illustrated in Fig~\ref{fig:system_Illustration}. Necessary symbols to describe the dynamics model are defined in the Tab.~\ref{tab:symbol definition}.  The payload is suspended on each aerial robot by a massless cable. It should be pointed out that frequent transitions of the cable between slack mode and taut mode will lead to switching of dynamics model, which is detrimental to control and can even cause dangerous system collapse. To address this risk, our planner will avoid the slack mode and generate a trajectory that always maintains the taut mode. Therefore, we only need to consider the following consistent dynamical model for the taut mode. The dynamics model of the payload is written as
\begin{subequations}
    \begin{align}
      \dot{\mathbf p}  &= \mathbf v, \\
      \label{equ:payload_model}  \dot{\mathbf v} &= - g \mathbf e_3 + \sum_{n = 1}^{N}  F_n \bm \rho_n /m_L.
    \end{align}
\end{subequations}
Besides, all the aerial robots share the same dynamics model and the dynamics model of the $n^{th}$ aerial robot is written as
\begin{subequations}
    \begin{align}
      \dot{\mathbf p}_n  &= \mathbf v_n, \\
      \label{eq:acceleration}  \dot{\mathbf v}_n &= - g \mathbf e_3 +  f_n \mathbf R_n \mathbf e_3 / m_n  - F_n \bm \rho_n /m_n, \\
    \dot{\mathbf R}_n &= \mathbf R_n \hat{\bm \omega}_n, \\ 
    \mathbf J_n \dot{\bm \omega}_n &= - \bm \omega_n \times \mathbf J_n \bm \omega_n  + \bm \tau_n,
    \end{align}
\end{subequations}
where $\hat \cdot$ is the operator to get the skew-symmetric matrix.

\subsection{Kinematic Constraint Elimination}
\label{subsec:Estimation model}
The $n^{th}$ taut cable imposes a kinematic constraint between the payload and the $n^{th}$ aerial robot
\begin{equation}
    \label{equ:kinematic_constraint}
    \|\mathbf p_n - \mathbf p \|_2 \equiv l.
\end{equation}

It should be pointed out that the differentials of Eq.~\ref{equ:kinematic_constraint} also implicitly enforce constraints on $\mathbf v_n, \mathbf v$ and their high-order derivatives. In this paper, we use payload's position $\mathbf p$ and $n^{th}$ cable's direction vector $\bm \rho_n$ to indirectly represent the $n^{th}$ aerial robot's position
\begin{equation}
    \label{equ:indirect representation}
    \mathbf p_n = \mathbf p + l \bm \rho_n,
\end{equation}
such that $\mathbf p, \mathbf p_n$, as well as their high-order derivatives always satisfy the complete kinematic constraint enforced by Eq.~\ref{equ:kinematic_constraint}, which is eliminated subsequently. 

In general, as in work \cite{sreenath2013dynamics}, the three components of force vector $F_n\bm\rho_n$ can be chosen as the variables to represent $F_n\bm\rho_n$. Thus, through unitizing $F_n\bm\rho_n$, $\bm\rho_n$ can be got to solve $\mathbf p_n$. However, when the norm $\|F_n \bm \rho_n\|_2$ converges to zero, there exists a singularity problem in solving $\mathbf p_n$. To avoid this singularity, we introduce two variables, i.e., the pitch angle $\theta_n$ and the azimuth angle $\phi_n$ of the $n^{th}$ cable as depicted in Fig.~\ref{fig:system_Illustration}. Thus, $\bm \rho_n$ can be represented as
\begin{equation}
    \label{equ:rhon}
    \bm \rho_n = \left[ \cos(\theta_n) \cos(\phi_n), \cos(\theta_n) \sin(\phi_n), \sin(\theta_n) \right]^T.
\end{equation} 

Let $ \bm{\xi}_n = (\theta_n, \phi_n, F_n)^T \in \mathbb{R}^{3}$, thus $F_n\bm\rho_n$ can be uniquely represented by $\bm{\xi}_n$. Using $ \bm{\xi}_n$ as the variable also brings an advantage that the constraints related to the direction and magnitude of $F_n\bm\rho_n$ can be enforced on $\bm \xi_n$ directly without constructing nonlinear maps to calculate $\bm \xi_n$ from $F_n\bm\rho_n$ such that these constraints can be eliminated via designing a diffeomorphism conveniently (Sec.~\ref{subsubsec:Topological Constraint Elimination}). 

\subsection{Extended Flat-output Variable}
\label{subsec:Extended_Flat_output_Variable}
The MARTS is a differential flatness system. For the $n^{th}$ aerial robot, $\mathbf p, \bm \xi_n$, and $\psi_n$ act as the flat-output variable. Compositing $\mathbf p$ and $\bm{\xi}_n, \psi_n $ for all $n \in [1, \cdots, N]$ yields a variable
\begin{equation}
    \mathbf Z = (\mathbf{p}^T, \bm{\xi}_1^T, \psi_1,  \cdots , \bm{\xi}_N^T, \psi_N)^T \in \mathbb{R}^{4N + 3},
\end{equation}
called as the extended flat-output variable for MARTS. Actually, $ \mathbf Z \backslash \{\bm \xi_N\}\in \mathbb{R}^{4N}$ can act as one selection of flat-output variable for MARTS since $ \mathbf Z \backslash \{\bm \xi_N\}$ is sufficient to uniformly determine $\bm \xi_N$ as elaborated in \cite{sreenath2013dynamics}. To avoid the complex map from $ \mathbf Z \backslash \{\bm \xi_N\}$ onto $\bm \xi_N$, we relieve $\bm \xi_N$'s binding on $ \mathbf Z \backslash \{\bm \xi_N\}$ and append redundant degree of freedom (DoF) $\bm \xi_N$ into $\mathbf Z$. As a result, we need to enforce a supererogatory dynamical constraint (Eq.~\ref{equ:payload_model}) onto $\mathbf Z$ during the optimization (Sec.~\ref{subsec:Coupling Dynamic Constraint}). The differential flatness characteristics help us to optimize a trajectory just on the low-dimensional extended flat-output space $\mathbf Z$. 

For any $s \in \{ s | s \in N_+, s \geq1\}$, we denote $\mathbf Z^{[s-1]} \in \mathbb{R}^{s\times(4N + 3)}$ as
\begin{equation}
    \mathbf Z^{[s-1]} = (\mathbf Z^T, \dot{\mathbf Z}^T,...,{\mathbf Z^{(s-1)}}^T)^T.
\end{equation}

\section{Planning for Safe and Agile Transportation}
\subsection{Flatness Maps}
\label{subsec:Flatness Maps}
In this section, we derive the flatness maps for the $n^{th}$ aerial robot based on the indirect representation (Eq.~\ref{equ:indirect representation}).  Firstly, from (Eq.~\ref{eq:acceleration}), we can get its mass-normalized thrust vector  
\begin{equation}
\label{eq:tau}
    \mathbf f_n = \ddot{\mathbf{p}} + l \ddot{\bm \rho}_n + g \mathbf e_3 + \mathrm F_n \bm {\rho}_n / m.
\end{equation}

Since the direction of $\mathbf f_n$ is parallel to $\mathbf {z}_{\mathcal B}^n$, we can obtain
\begin{equation}
\label{eq:zb}
    \mathbf{z}_{\mathcal{B}}^{n} = \mathcal N ({\mathbf f_n}), 
\end{equation}
where $\mathcal N(): \mathbb R^3 \to \mathbb R^3 $ is the vector unitization function defined as $ \mathcal N(\mathbf x) \triangleq \mathbf  x / \norm{\mathbf x}_2, \mathbf x\in \mathbb R^3  $. To recover the quaternion $\mathbf q_n$ w.r.t $\mathbf R_n$ from $\mathbf{z}_{\mathcal B}^n, \psi_n$, we use the rotation factorization known as Hopf Fibration \cite{watterson2019control}, which introduces the fewest singularities. The body frame $\mathbf{z}_{\mathcal B}^n$ is constructed through rotating the world frame $\mathcal I$ around its z-axis $\mathbf{z}_{\mathcal I}$ by $\psi_n$ and continually rotating it around the current y-axis till its z-axis coincides with $\mathbf z_{\mathcal B}^n$, whose quaternion $\mathbf q_n$ can be computed as 
\begin{align}
    \mathbf {q}_n =\mathbf {q}_{\mathbf z_{\mathcal B}^{n}} \odot \mathbf {q}_{\psi_n}, 
\end{align}
where $\odot$ is the quaternion multiplication. For arbitrary vector $\mathbf x \in \mathbb R^3 $ and quaternion $\mathbf q$, let $\mathbf x_{1}, \mathbf x_{2}, \mathbf x_{3}$ denote the three elements of $\mathbf x$, and let $\mathbf q^w, \mathbf q^x, \mathbf q^y, \mathbf q^z $ denote the four elements of $\mathbf q$. Thus, $\mathbf{q}_\psi^i, \mathbf{q}_{\mathbf z_{\mathcal B}^{n}}$ can be expanded as 
\begin{subequations}
\begin{align}
    \mathbf{q}_{\psi_n} &=\left(\cos \frac{\psi_n}{2}, 0,0, \sin \frac{\psi_n}{2}\right), \\ 
  \mathbf{q}_{\mathbf z_\mathcal{B}^n} &=\frac{1}{\sqrt{2(\mathbf z_{\mathcal B,3}^n+1)}}\left(\mathbf z_{\mathcal B,3}^n+1, -\mathbf z_{\mathcal B,2}^n,\mathbf z_{\mathcal B,1}^n ,0\right).
\end{align}
\end{subequations}
 
Since the tile angle $\vartheta_n$ is only related to $\mathbf q_{n}^x, \mathbf q_{n}^y$, we just give the formulation of $\mathbf {q}_n^x, \mathbf {q}_n^y$ as follows, 
\begin{subequations}
\begin{align}
\mathbf{q}^x_n &=\frac{-1}{\sqrt{2(\mathbf z_{\mathcal B,3}^n+1)}} \left(\mathbf z_{\mathcal B,2}^n \cos \frac{\psi_n}{2} - \mathbf z_{\mathcal B,1}^n \sin \frac{\psi_n}{2}\right), \\ 
\mathbf{q}^y_n &=\frac{1}{\sqrt{2(\mathbf z_{\mathcal B,3}^n+1)}} \left(\mathbf z_{\mathcal B,1}^n \cos \frac{\psi_n}{2} + \mathbf z_{\mathcal B,2}^n \sin \frac{\psi_n}{2}\right).
\end{align}  
\end{subequations}

Now, the tilt angle $\vartheta_{n}$ can be calculated by 
\begin{align}
    \vartheta_{n} = \text{acos}\left[1 - 2 \left({ \mathbf q^x_n}^2 + { \mathbf q^y_n}^2\right)\right],
\end{align}
and the body rate $\bm \omega_n$ can be calculated by 
\begin{align}
    \bm \omega_n = 2 \mathbf {q}_{n}^{-1} \odot \dot{\mathbf{q}}_n. 
\end{align}

Finally, inverting and differentiating $\mathbf q_n$ gives
\begin{subequations}
\begin{align}
    \label{equ:omega1}  \bm \omega_{n,1} =& \dot{\mathbf z}_{\mathcal B,1}^n \sin (\psi_n) - \dot{\mathbf z}_{\mathcal B,2}^n \cos (\psi_n) \notag \\ 
     & - \frac{ \dot{\mathbf z}_{\mathcal B,3}^n \left[\mathbf z_{\mathcal B,1}^n \sin (\psi_n) - \mathbf z_{\mathcal B,2}^n \cos (\psi_n)\right]}{\mathbf z_{\mathcal B,3}^n+1}, \\ 
    \label{equ:omega2} \bm\omega_{n,2} =& \dot{\mathbf z}_{\mathcal B,1}^n \cos (\psi_n) + \dot{\mathbf z}_{\mathcal B,2}^n \sin (\psi_n) \notag \\ 
    & - \frac{ \dot{\mathbf z}_{\mathcal B,3}^n \left[\mathbf z_{\mathcal B,1}^n \cos (\psi_n) + \mathbf z_{\mathcal B,2}^n \sin (\psi_n)\right]}{\mathbf z_{\mathcal B,3}^n+1}, \\
    \label{equ:omega3} \bm \omega_{n,3} =& \frac{\mathbf z_{\mathcal B,2}^n \dot{\mathbf z}_{\mathcal B,1}^n - \mathbf z_{\mathcal B,1}^n \dot{\mathbf z}_{\mathcal B,2}^n}{\mathbf z_{\mathcal B,3}^n+1} + \dot{\psi}_n, 
\end{align}  
\end{subequations}
where $\dot{\mathbf{z}}_{\mathcal B}^n$ is the vector differentiation derived as 
\begin{subequations}
    \begin{align}
    \label{eq:dzb} \dot{\mathbf{z}}_{\mathcal B}^n &= \frac{1}{\| \mathbf f_n \|_2} \left( {\mathbf{I}} - \frac{\mathbf f_n \mathbf f_n^T }{\|\mathbf f_n\|_2^2} \right) \dot{\mathbf f}_n, \\
    \dot{\mathbf f}_n &= \dddot{\mathbf{p}} + l\dddot{\bm \rho}_n + \frac{1}{m} \left( \dot{ F}_n \bm {\rho}_n + F_n  \dot{\bm {\rho}}_n \right).
    \end{align}
\end{subequations}

\subsection{Trajectory Representation}
\label{subsec:traj rep}
Flatness maps rely only on the flat-output variable, which avoid integrating the dynamical equations to obtain the state and the control input. In this paper, we use a multidimensional piecewise polynomial to represent the flat-output trajectory so that the motion planning for MARTS can be performed by the deformation of the flat-output trajectory. Let $\mathbf Z(t)$ denote a $(4N+3 )$-dimension, $M$-piece polynomial of $2s$-order. The $m^{th}$ piece of $\mathbf Z(t)$ is defined as   
\begin{equation}
    \label{equ:m-th piece}
    \mathbf Z|_m(t)=\mathbf c_m^T \bm \beta(t),~~\forall{t}\in[0,T_m],
\end{equation}
where $\mathbf c_m \in \mathbb{R}^{2s\times (4N +3)}$ is the coefficient matrix, $\bm \beta(t)=[1,t^1,\cdots,t^{2s - 1}]^T$ is the natural basis and $T_m \in  \mathbb{R}_{>0}$ is the time duration of the $m^{th}$ piece. Thus, the flat-output trajectory $\mathbf Z(t) $ can be fully represented by the coefficient matrix $\mathbf {c}=(\mathbf c_1^T,\cdots,\mathbf c_M^T)^T\in \mathbb{R}^{2sM \times (4N+3)}$ and the time vector $ \mathbf{T}=(T_1,\cdots,T_M)^T\in \mathbb{R}^M_{>0}$. 

To improve the efficiency of trajectory optimization, we adopt $\mathcal C_{\mathbf{MINCO}}$ \cite{9765821}, a state-of-the-art (SOTA) polynomial trajectory class. Given a set of sparse representation parameters $\{\mathbf w, \mathbf T\}$, $\mathcal C_{\mathbf{MINCO}}$ constructs a system of linear equations $\mathcal L$ to solve $\{\mathbf c, \mathbf T\}$ from $\{\mathbf w, \mathbf T\}$ 
\begin{equation}
    \label{equ:MCB}
    (\mathbf c, \mathbf T)= \mathcal L(\mathbf w, \mathbf T), 
\end{equation}
such that a piecewise polynomial, i.e., the flat-output trajectory $\mathbf Z(t)$ is generated, where $\mathbf w = (\mathbf w_1, \cdots, \mathbf w_{M-1})^T \in \mathbb{R}^{(4N+3)\times (M - 1)}$ and for any $m' \in [1, \cdots, M-1]^T$, $\mathbf w_{m'} = ({\mathbf{p}^{m'}}^T, {\bm{\xi}_1^{m'}}^T, \psi_1^{m'},  \cdots , {\bm{\xi}_N^{m'}}^T, \psi_N^{m'})^T \in \mathbb{R}^{4N+3}$ is the ${m'}^{th}$ junction to connect the $ {m'}^{th}$ piece and the $ (m'+1)^{th}$ piece. Using $\{\mathbf w,\mathbf T\}$ to determine the continuity conditions at the junctions, with the boundary conditions at the initial and final moments, $\mathcal C_{\mathbf{MINCO}}$ can uniquely generate a minimum control effort trajectory that satisfies all these conditions by solving $\mathcal L$ directly, while avoiding time-consuming iterative optimization.

\subsection{Safety Constraints}
\label{subsec:Safety Constraints}
The collapse of MARTS occurs when any part of the system collides with the obstacles in a complex environment, or when any two aerial robots within the system reciprocally collide with each other. In this section, we construct safety constraints to avoid obstacle collisions and reciprocal collisions between the aerial robots. Besides, we also design the differential metrics and derive corresponding gradients w.r.t trajectory's parameter, i.e., ${\mathbf c, \mathbf T}$. In the following sections, we assume that all the states and control input defined in Tab.~\ref{tab:symbol definition} are associated with a special point on the $m^{th}$ piece of trajectory. The details are given as follows.

\subsubsection{Obstacle Avoidance Constraint}
The obstacle avoidance constraint is designed to avoid obstacle collisions. Complete obstacle avoidance means that the $N$ surfaces swept by all cables' trajectories, the $N$ aerial robots' trajectories, and the payload's trajectory should not be intersected by any obstacle. In this work, we maintain an Euclidean Signed Distance Field (ESDF). To evaluate the safety margin between a point $\mathbf p^* \in \mathbb R^3$ and the obstacle, we design a differential metric $\mathcal{M}_{oa}: \mathbb R^3  \times \mathbb R_{>0} \to \mathbb R$ as 
\begin{align}
    \mathcal{M}_{oa}(\mathbf{p}^*,d) =  d - \mathcal{E} \left(\mathbf p^* \right), 
\end{align}
where $\mathcal{E} \left(\mathbf p^* \right) $ is the distance between $\mathbf p^*$ and its closest obstacle evaluated by ESDF and $d \in \mathbb R_{>0}$ is the safety distance away from obstacle. 

Let $d_{oa}^c \in \mathbb R_{>0}, d_{oa}^Q  \in \mathbb R_{>0}, d_{oa}^L  \in \mathbb R_{>0} $ denote the safe distances of the cable's sampled point, the aerial robot, and the payload respectively. We enforce the obstacle avoidance constraint on the payload
\begin{align}
    \label{equ:obstacle_avoidance_constraint3} 
    \mathcal{M}_{oa}(\mathbf p, d_{oa}^L ) < 0.    
\end{align}
For any $ n \in [1, \cdots, N]$, we enforce obstacle avoidance constraint on the $n^{th}$ aerial robot 
\begin{align}
    \label{equ:obstacle_avoidance_constraint2} 
    \mathcal{M}_{oa}(\mathbf p_n, d_{oa}^Q ) <0.
\end{align} 
For the $n^{th}$ cable, we uniformly sample $K$ points along the cable. For each $k\in [1, \cdots, K]$ and each $ n \in [1, \cdots, N]$, we enforce the obstacle avoidance constraint on the $k^{th}$ sampled points $\mathbf{p}_{n}^{k}$ along the $n^{th}$ cable
\begin{align}
    \label{equ:obstacle_avoidance_constraint1}
    \mathcal{M}_{oa}(\mathbf p_n^k, d_{oa}^c) < 0,
\end{align}
where $\mathbf{p}_{n}^{k}$ can be computed as 
\begin{align}
    % \label{equ:obstacle_avoidance_constraint}
    \mathbf p_n^k = \mathbf p + \frac{lk}{K + 1} \bm{\rho_{n}}.
\end{align}

Then the penalty function $\mathcal J_{oa}^{0}$ for the constraint enforced on the payload in Eq.~\ref{equ:obstacle_avoidance_constraint3} is defined as 
\begin{align}
    \label{load obstacle avoidance cost function}
      \mathcal J_{oa}^{0}  =  \mathcal L_\mu \left[\mathcal{M}_{oa}(\mathbf p, d_{oa}^L ) \right],
\end{align}
Besides, the penalty function $\mathcal J_{oa}^{n}$ for the constraints w.r.t both the $n^{th}$ aerial robot and $n^{th}$ cable in Eq.~\ref{equ:obstacle_avoidance_constraint2} - Eq.~\ref{equ:obstacle_avoidance_constraint1} is defined as 
\begin{align}
    \label{cable obstacle avoidance cost function}
      \mathcal J_{oa}^{n}  = &  \sum_{k = 1}^{K}  \mathcal L_\mu \left[ \mathcal{M}_{oa}(\mathbf p_n^k, d_{oa}^c) \right]  + \mathcal L_\mu \left[\mathcal{M}_{oa}(\mathbf p_n, d_{oa}^Q ) \right],
\end{align}
where $\mathcal L_\mu : \mathbb R \to \mathbb R $ is a $C^2$-smoothing function defined as 
    \begin{align}
      \mathcal L_\mu(x) = 
      \begin{cases}
        0,                    & x \leq 0,       \\
        (\mu - x/2)(x/\mu)^3, & 0 < x \leq \mu, \\
        x - \mu/2,            & x > \mu,
      \end{cases}
    \end{align}
which smooths the transition of a truncated linear function at $x= 0$.

 Finally, the gradients of $\mathcal{M}_{oa}(\mathbf p, d_{oa}^L )$ w.r.t ${\mathbf c_{m}, T_m}$ are derived as 
\begin{subequations}
    \begin{align}
        \frac{\partial \mathcal M_{oa}(\mathbf p, d_{oa}^L)}{\partial \bm c_{m,{\mathbf p}}}  =& -{\nabla \mathcal E(\mathbf p)}^T \otimes \bm \beta , \\ 
        \frac{\partial \mathcal M_{oa}(\mathbf p, d_{oa}^L)}{\partial T_m}  =& -  {\nabla \mathcal E(\mathbf p)}^T \mathbf v,
     \end{align}
    \label{equ:thrust partial derivative}
\end{subequations}
where $ \otimes $ is the Kronecker product and $\nabla\mathcal E: \mathbb R^3 \to \mathbb R^3$ is an operator to query the ESDF gradient.

Let $ \bm \Phi_n$ denotes $(\theta_n, \phi_n)^T $ and $\varrho_k = {lk}/ {(K + 1)}$. The gradients of $\mathcal{M}_{oa}(\mathbf p_n^k, d_{oa}^c )$ w.r.t ${\mathbf c_m, T_m}$ are computed as 
\begin{subequations}
    
    \begin{align}
        \label{equ:obstacle_avoidance_gradient1}
        \frac{\partial \mathcal M_{oa}(\mathbf p_n^k, d_{oa}^c)}{\partial \mathbf c_{m,\mathbf p}}  =& -{\nabla \mathcal E(\mathbf p_n^k)}^T \otimes \bm \beta , \\ 
        \frac{\partial \mathcal M_{oa}(\mathbf p_n^k, d_{oa}^c)}{\partial \mathbf c_{m, \bm \Phi_n}}  =& -\varrho_k\left(  \nabla \mathcal E(\mathbf p_n^k)^T  \frac{\partial \bm \rho_n}{\partial \bm \Phi_n} \right) \otimes \bm \beta , \\
        \frac{\partial \mathcal M_{oa}(\mathbf p_n^k, d_{oa}^c)}{\partial \mathbf c_{m, F_n}}  =& \bm 0  ,\\ 
        \label{equ:obstacle_avoidance_gradient4}
        \frac{\partial \mathcal M_{oa}(\mathbf p_n^k, d_{oa}^c)}{\partial T_m}  =& -  \left( \dot{\mathbf p} + \varrho_k \frac{\partial \bm \rho_n}{\partial \bm \Phi_n} \dot{\bm \Phi}_n \right)^T {\nabla \mathcal E(\mathbf p_n^k)}.
     \end{align}
\end{subequations}

Note that $\mathbf c_{m, \mathbf p}$ denotes the submatrix of $ \mathbf c_m$ w.r.t the variable $\mathbf p$ in $\mathbf Z$. The gradients of $\mathcal{M}_{oa}(\mathbf p_n, d_{oa}^Q )$ w.r.t ${\mathbf c_m, T_m}$ can be computed similarly as in Eq.~\ref{equ:obstacle_avoidance_gradient1} - Eq.~\ref{equ:obstacle_avoidance_gradient4} except that the $\mathbf p_n^k, d_{oa}^c $, and $ \varrho_k $ are need to be substituted by $ \mathbf p_n, d_{oa}^Q$, and $1$, which are omitted here.

\subsubsection{Reciprocal Avoidance Constraint}
The reciprocal avoidance constraint is designed to avoid reciprocal collisions among the aerial robots. To evaluate the safety margin between two aerial robots, we design a differential metric $\mathcal{M}_{ra}: \mathbb R^3  \times  \mathbb R^3  \times  \mathbb R_{>0} \to \mathbb R$ as 
\begin{equation}
    \label{equ:reciprocal_avoidance_constraint}
    \mathcal{M}_{ra}(\mathbf{p},\mathbf{p}', d) =  {d}^2 -  \left\| \mathbf{p} - \mathbf{p}'  \right\|^2,
\end{equation}
For each $ n\in [1, \cdots, N - 1]$ and each $  n' \in [n + 1, \cdots, N]$, we enforce the reciprocal avoidance constraint between the $n^{th}$ aerial robot and the $n'^{th}$ aerial robot
\begin{subequations}
    \begin{align}
        \label{equ:reciprocal_avoidance_constraint}
          &\mathcal{M}_{ra}(\mathbf p_n, \mathbf p_{n'},  d_{ra}^Q) < 0,
    \end{align}
\end{subequations}
where $d_{ra}^Q \in \mathbb R_{>0}$ is the safe distance between the aerial robots. Then, the penalty function $\mathcal J_{ra}$ for all the reciprocal avoidance constraints is defined as
\begin{align}
    \label{quadrotor reciprocal avoidance cost function}
      \mathcal J_{ra}  = &  \sum_{i = 1}^{N-1} \sum_{j = i + 1}^{N}  \mathcal L_\mu \left[ \mathcal{M}_{ra}(\mathbf p_i, \mathbf p_j, d_{ra}^Q) \right].
\end{align}
Finally, the gradients of $\mathcal M_{ra}(\mathbf p_i, \mathbf p_j, d_{ra}^Q) $ w.r.t ${\mathbf c_m, T_m}$ are computed as 
\begin{subequations}
    \label{equ:thrust partial derivative}
    \begin{align}
        \frac{\partial \mathcal M_{ra}(\mathbf p_i, \mathbf p_j,d_{ra}^Q)}{\partial \mathbf c_{m,\mathbf p}}  =& \bm 0 ,   \\ 
        \frac{\partial \mathcal M_{ra}(\mathbf p_i, \mathbf p_j,d_{ra}^Q)}{\partial \mathbf c_{m, \bm \rho_i}}  =& - 2l\left[{(\mathbf{p}_i - \mathbf{p}_j)}^T \frac{\partial \bm \rho_i}{\partial \bm \Phi_i} \right] \otimes \bm \beta ,  \\
        \frac{\partial \mathcal M_{ra}(\mathbf p_i, \mathbf p_j,d_{ra}^Q)}{\partial \mathbf c_{m,\bm \rho_j}}  =&  2l\left[{(\mathbf{p}_i - \mathbf{p}_j)}^T \frac{\partial \bm \rho_j}{\partial \bm \Phi_j} \right] \otimes \bm \beta , \\
        \frac{\partial \mathcal M_{ra}(\mathbf p_i, \mathbf p_j,d_{ra}^Q)}{\partial \mathbf c_{m, {F}_i}}  =&  \frac{\partial \mathcal M_{ra}(\mathbf p_i, \mathbf p_j,d_{ra}^Q)}{\partial \mathbf c_{m, {F}_j}}  =  \bm 0 , \\ 
        \frac{\partial \mathcal M_{ra}(\mathbf p_i, \mathbf p_j,d_{ra}^Q)}{\partial T_m}  =& - 2l\left(\frac{\partial \bm \rho_i}{\partial \bm \Phi_i} \dot{\bm \Phi}_i - \frac{\partial \bm \rho_j}{\partial \bm \Phi_j} \dot{\bm \Phi}_j \right)^T \notag \\ & ~~~~~~ \cdot {(\mathbf{p}_i - \mathbf{p}_j)} . 
     \end{align}
\end{subequations}

% Enhancing the maneuverability while ensuring safety is essential to improve the efficiency of PT.  Based on the constraints, we also construct the cost functions and derive corresponding gradients w.r.t trajecotry parameters, i.e., ${\bm c, \bm T}$. 

\subsection{Dynamical Feasibility Constraints}
\label{subsec:Dynamical Feasibility Constraints}
For any aerial robot in the MARTS, an overrun of either its state or control input will cause the MARTS to crash. In this section, we enforce dynamical feasibility constraints on the state and control input of each aerial robot to ensure that an agile trajectory can be executed by the practical MARTS. These constraints are designed to limit transient state and control input along the trajectory when the agility enhances as the trajectory's duration decreases, which involve four types for each aerial robot, i.e., a constraint on maximum velocity, a constraint on maximum and minimum mass-normalized thrust, a constraint on maximum tilt angle, and a constraint on maximum body rate. The details are given as follows.
\subsubsection{Maximum Velocity Constraint}
For a scalar $a$, we define a metric $\mathcal{M} : \mathbb R  \times \mathbb R_{>0} \to \mathbb R$ to evaluate whether $a$ exceed its bound $\bar{a}$
\begin{align}
    \mathcal{M}(a,\bar{a}) =  a - \bar{a}. 
\end{align}
Thus, for each $n \in [1, \cdots, N]$, we constrain the maximum velocity for the $n^{th}$ aerial robot
\begin{equation}
        \label{equ:maximum_velicity1}
          \left\|\mathbf v_n\right\|^2 < v_{max}^2,          
\end{equation}
where $v_{max}$ is the maximum allowable velocity of aerial robot. Then the penalty function for the constraint w.r.t in Eq.~\ref{equ:maximum_velicity1} is defined as
\begin{equation}
    \label{max velocity cost function}
      \mathcal J_{\mathbf v}^n  = \mathcal L_\mu \left( \mathcal{M}_{df}^{\mathbf v_n}\left(\left\|\mathbf v_n\right\|^2,{v_{max}^2}\right) \right).
\end{equation}
% \begin{subequations}
%     \begin{align}
%         \frac{\partial \mathcal J_{v}^0}{\partial \bm c_{\mathbf p}}  =&  \left( \frac{\partial \mathcal J_{v}^0}{\partial \mathbf v} \right)^T  \otimes \dot{\bm \beta},  \\ 
%         \frac{\partial \mathcal J_{v}^0}{\partial T_m}  =& \left( \frac{\partial \mathcal J_{v}^0}{\partial \mathbf v} \right)^T \mathbf a.
%     \end{align}
%     \label{equ:thrust partial derivative}
% \end{subequations}
In Eq.~\ref{max velocity cost function}, a superscript $\mathbf v_n$ and a subscript $df$ is added to the symbol $\mathcal{M}$ to denote the dynamical constraint w.r.t this superscript. In the following of this paper, $\mathcal{M}_{df}^{\mathbf v_n} \left(\left\|\mathbf v_n\right\|^2,{v_{max}^2}\right) $ will be denoted by $\mathcal{M}_{df}^{\mathbf v_n}$ for the sake of simplification. Finally, the gradients of $\mathcal{M}_{df}^{\mathbf v_n}$ w.r.t ${\mathbf c_m, T_m}$ can be computed as 
\begin{subequations}
    \label{equ:thrust partial derivative}
    \begin{align}
        \frac{\partial \mathcal{M}_{df}^{\mathbf v_n}}{\partial \mathbf c_{m,\mathbf p}}  =&  \left( \frac{\partial \mathcal{M}_{df}^{\mathbf v_n}}{\partial \bm \dot{\mathbf p}} \right)^T  \otimes \dot{\bm \beta},  \\ 
        \frac{\partial \mathcal{M}_{df}^{\mathbf v_n}}{\partial \mathbf c_{m, \bm{\rho}_n}} = & \sum_{j \in \Pi_{\mathbf v}} \left[ \left( \frac{\partial \mathcal{M}_{df}^{\mathbf v_n}}{\partial \dot{\bm \rho}_n} \right)^T \frac{\partial \dot{\bm \rho}_{n}}{\partial \bm \Phi^{(j)}_n} \right] \otimes \bm \beta^{(j)},  \\
        \frac{\partial \mathcal{M}_{df}^{\mathbf v_n}}{\partial \mathbf c_{m, F_n}} =& \bm 0,  \\ 
        \frac{\partial \mathcal{M}_{df}^{\mathbf v_n}}{\partial T_m}  =& \left( \frac{\partial \mathcal{M}_{df}^{\mathbf v_n}}{\partial \dot{\bm \rho}_n} \right)^T \sum_{j \in \Pi_{\mathbf v}} \frac{\partial \dot{\bm \rho}_{n}}{\partial \bm \Phi^{(j)}_n} \bm \Phi^{(j + 1)}_n  \notag \\
        &+\left( \frac{\partial \mathcal{M}_{df}^{\mathbf v_n}}{\partial \bm \dot{\mathbf p}} \right)^T \ddot{\mathbf p},
    \end{align}
\end{subequations}
where $\Pi_{\mathbf v} = \{0, 1\}$ is an index set related to this constraint. 
\subsubsection{Maximum and Minimum Thrust Constraint}
A realistic aerial robot can only provide a limited thrust. Therefore, to ensure the system controllability, for all $n \in [1, \cdots, N]$, we restrict the mass-normalized thrust $\|\mathbf f_n\| $ to be less than the maximum thrust $f_{max}$ that can be provided by a practical aerial robot and more than the minimum thrust $f_{min} $
\begin{align}
    \label{equ:net_thrust_constraint}
    f_{min} \leq \norm{\mathbf f_n} \leq f_{max}.  
\end{align}
Let $f_{avg}$ denote $ (f_{max} + f_{min}) / 2 $, and $f_{rag}$ denote $ (f_{max} - f_{min}) / 2 $, the penalty function for the constraint in Eq.~\ref{equ:net_thrust_constraint} is defined as 
\begin{align}
    \label{tau cost function}
      \mathcal J_{\mathbf f}^n  = & \mathcal L_\mu \left(  \mathcal{M}_{df}^{\mathbf f_n}\left(  \left( \|\mathbf f_n \| -  f_{avg}  \right)  ^2,  f_{rag} \right) \right). 
\end{align}
Then the gradients of $\mathcal M_{df}^{\mathbf f_n } $ w.r.t ${\mathbf c_m, T_m}$ can be calculated as  
\begin{subequations}
    \label{equ:thrust partial derivative}
    \begin{align}
        \label{equ:grad_tau_c} \frac{\partial \mathcal M^{\mathbf f_n}_{df}}{\partial \mathbf c_{m,\mathbf p}}  =&  \left[ \left( \frac{\partial \mathcal M^{\mathbf f_n}_{df}}{\partial \mathbf f_n} \right)^T \frac{\partial \mathbf f_n}{\partial  \ddot{\mathbf p}} \right] \otimes \ddot{\bm \beta},  \\ 
        \frac{\partial \mathcal{M}_{df}^{\mathbf f_n}}{\partial \mathbf c_{m, \bm \rho_n}}  =&  \sum_{i \in \Pi_{\mathbf f}} \sum_{j=0}^{i}  \left[ \left( \frac{\partial \mathcal{M}_{df}^{\mathbf f_n}}{\partial \mathbf f_n} \right)^T \frac{\partial \mathbf f_n}{\partial \bm \rho^{(i)}_n}\frac{\partial \bm \rho^{(i)}_n}{\partial \bm \Phi^{(j)}_n} \right] \otimes \bm \beta^{(j)} , \\
        \frac{\partial \mathcal \mathcal M^{\mathbf f_n}_{df}}{\partial \mathbf c_{m,  F_n}} =& \left[ \left( \frac{\partial \mathcal M^{\mathbf f_n}_{df}}{\partial \mathbf f_n} \right)^T \frac{\partial \mathbf f_n}{\partial  F_n} \right] \bm \beta,  \\ 
        \label{equ:grad_tau_t} \frac{\partial \mathcal{M}_{df}^{\mathbf f_n}}{\partial T_m}  =&\sum_{i \in \Pi_{\mathbf f}} \sum_{j=0}^{i}  \left( \frac{\partial \mathcal{M}_{df}^{\mathbf f_n}}{\partial \mathbf f_n} \right)^T \frac{\partial \mathbf f_n}{\partial \bm \rho^{(i)}_n}\frac{\partial \bm \rho^{(i)}_n}{\partial \bm \Phi^{(j)}_n} \bm \Phi^{(j + 1)}_n  \notag \\
        & +\left( \frac{\partial \mathcal{M}_{df}^{\mathbf f_n}}{\partial \mathbf f_n} \right)^T  \left( \frac{\partial \mathbf f_n}{\partial  \ddot{\mathbf p}}  \dddot{\mathbf p}  +  \frac{\partial \mathbf f_n}{\partial F_n} \dot{F}_n \right).
    \end{align}
\end{subequations}
where the index set $\Pi_{\mathbf f} = \{0, 2\}$. 
\subsubsection{Maximum Tilt Angle Constraints}
For each $n \in [1, \cdots, N]$, we limit the maximum tilt angle $\vartheta_n$ for the $n^{th}$ aerial robot
%to ensure the control precision 
\begin{align}
    \label{equ:tilt_angle_constraint}
    \vartheta_n \leq \vartheta_{max},
\end{align}
where $\vartheta_{max}$ is the maximum admissible tilt angle. Thus, the penalty function for the constraint in Eq.~\ref{equ:tilt_angle_constraint} can be defined as
\begin{align}
    \label{equ:cost_tile_angle}
      \mathcal J_{\vartheta}^n = \mathcal L_\mu \left( \mathcal M_{df}^{\vartheta_n} \left(\vartheta_n,  \vartheta_{max}\right)\right). 
\end{align}
Since $\mathcal M_{df}^{\vartheta_n}$ relies on $\mathbf f_n$ as $\mathcal M_{df}^{\mathbf f_n } $, the formulation of $\frac{\partial \mathcal M_{df}^{\vartheta_n} }{\partial \mathbf c_m} \frac{\partial \mathcal \mathcal M_{df}^{\vartheta_n}}{\partial T_m}$ is similar to Eq.~\ref{equ:grad_tau_c} - Eq.~\ref{equ:grad_tau_t} except that $\mathcal M_{df}^{\mathbf f_n } $ need to be substituted by $\mathcal M_{df}^{\vartheta_n}$ and $\frac{\partial \mathcal M_{df}^{\vartheta_n}}{\partial \mathbf f_n} $ can be computed by
\begin{align}
    \frac{\partial \mathcal M^{\vartheta_n}_{df}}{\partial \mathbf f_n} =  \sum_{i \in \Pi_{\vartheta}}  \frac{\partial \mathcal \mathcal M^{\vartheta_n}_{df}}{\partial  q^i_n} \left(\frac{\partial q^i_n}{\partial \mathbf z_{\mathcal B}^n} \right)^T \frac{\mathbf z_{\mathcal B}^n}{\partial \mathbf f_n},
\end{align}
where the index set $\Pi_{\vartheta} = \{x, y\}$. 
\subsubsection{Maximum Body Rate Constraint}
For each $n \in [1, \cdots, N]$, we restrict the upper bound of body rate
\begin{align}
    \label{equ:body_rate_constraint}
     \norm{\bm \omega_n} \leq \omega_{max},
\end{align}
where $\omega_{max}$ is the maximum admissible body rate for aggressive flight. The penalty function for the constraint in Eq.~\ref{equ:body_rate_constraint} is defined as
\begin{align}
    \label{equ:cost_body_rate}
      \mathcal J_{\bm \omega}^n  = \mathcal L_\mu  \left( \mathcal M_{df}^{\bm \omega_n} \left(\norm{\bm \omega_n}^2 ,  \omega_{max}^2\right)\right).
\end{align}
And the gradients of $\mathcal M_{df}^{\bm \omega_n} $ w.r.t ${\mathbf c_m, T_m}$ are computed as 
\begin{subequations}
    \label{equ:thrust partial derivative}
    \begin{align}
        \frac{\partial \mathcal M_{df}^{\bm \omega_n}}{\partial \mathbf c_{m, \mathbf p}}  =&\sum_{i \in \Pi_{\bm \omega}} \left[\left( \frac{\partial \mathcal M_{df}^{\bm \omega_n}}{\partial \mathbf f_n^{(i)}} \right)^T \frac{\partial \mathbf f_n^{(i)}}{\partial  \mathbf p^{(i + 2)}} \right] \otimes \bm \beta^{(i + 2)}, \\ 
        \frac{\partial \mathcal M_{df}^{\bm \omega_n}}{\partial \mathbf  c_{m, \bm \rho_n}}  =& \sum_{i \in \Pi_{\bm \omega}}  \sum_{j \in \Pi_{\bm \omega}^{i}} \sum_{k=0}^{j}  \left[ \left( \frac{\partial \mathcal M_{df}^{\bm \omega_n}}{\partial \mathbf f_n^{(i)}} \right)^T \frac{\partial \mathbf f_n^{(i)}}{\partial \bm \rho^{(j)}_n}\frac{\partial \bm \rho^{(j)}_n}{\partial \bm \Phi^{(k)}_n} \right]   \notag \\
        & \ \ \ \ \ \  \ \ \ \ \ \  \ \ \ \ \  \otimes \bm \beta^{(k)} , \\
        \frac{\partial \mathcal M_{df}^{\bm \omega_n}}{\partial \mathbf  c_{m, F_n}} =& \sum_{i \in \Pi_{\bm \omega}} \sum_{j = 0}^{i}  \left[ \left( \frac{\partial \mathcal M_{df}^{\bm \omega_n}}{\partial \mathbf f_n^{(i)}} \right)^T \frac{\partial \mathbf f_n^{(i)}}{\partial  F^{(j)}_n} \right] \bm \beta^{(j)},  \\
        \frac{\partial \mathcal M_{df}^{\bm \omega_n}}{\partial T_m}  = & \sum_{i \in \Pi_{\bm \omega}}  \sum_{j \in \Pi_{\bm \omega}^i} \sum_{k=0}^{j}  \left( \frac{\partial \mathcal M_{df}^{\bm \omega_n}}{\partial \mathbf f_n^{(i)}} \right)^T \frac{\partial \mathbf f_n^{(i)}}{\partial \bm \rho^{(j)}_n}\frac{\partial \bm \rho^{(j)}_n}{\partial \bm \Phi^{(k)}_n} \bm \Phi^{(k + 1)}_n \notag \\ 
        & + \sum_{i \in \Pi_{\bm \omega}} \sum_{j = 0}^{i}  \left( \frac{\partial \mathcal M_{df}^{\bm \omega_n}}{\partial \mathbf f_n^{(i)}} \right)^T 
        \frac{\partial \mathbf f_n^{(i)}}{\partial  F^{(j)}_n} F^{(j + 1)}_n \notag \\
        & + \sum_{i \in \Pi_{\bm \omega}} \left( \frac{\partial \mathcal M_{df}^{\bm \omega_n}}{\partial \mathbf f_n^{(i)}} \right)^T \frac{\partial \mathbf f_n^{(i)}}{\partial  \mathbf p^{(i + 2)}} \mathbf p^{(i + 3)},  
    \end{align}
\end{subequations}
where the index sets are defined respectively as $\Pi_{\bm \omega} = \{0, 1\}, \Pi_{\bm \omega}^{0} = \{0, 2\}, \Pi_{\bm \omega}^{1} = \{0, 1, 3\}$. Besides, for any $i \in \Pi_{\bm \omega}$, $\frac{\partial \mathcal M_{df}^{\bm \omega_n}}{\partial \mathbf f_n^{(i)}} $ can be further calculated as 
\begin{align}
    \frac{\partial \mathcal M_{df}^{\bm \omega_n}}{\partial \mathbf f_n^{(i)}} =  \sum_{j  = i}^{1}  \left( \frac{\partial \mathcal M_{df}^{\bm \omega_n}}{\partial  \bm \omega_n} \right)^T \frac{\partial \bm \omega_n}{\partial {\mathbf{z}_{\mathcal B}^n}^{(j)} }  \frac{{\mathbf{z}_{\mathcal B}^n}^{(j)}}{\partial {\mathbf f_n^{(i)}}}.
\end{align}

\subsection{Coupling Dynamic Constraint}
\label{subsec:Coupling Dynamic Constraint}
As has exposited in Sec.~\ref{subsec:Extended_Flat_output_Variable}, we need to enforce an extra payload dynamical constraint on the extended flat-output variable $\mathbf Z$ based on Eq.~\ref{equ:payload_model}
\begin{align}
    \label{equ:payload_dynamical_constraint}
    \dot{\mathbf v} \equiv - g \mathbf e_3 + \sum_{n = 1}^{N}  F_n \bm \rho_n /m_L.
\end{align}
Then we construct the penalty function for this constraint
\begin{align}
    \label{equ:payload_dynamical_penalty}
    \mathcal J_d =  \mathcal M \left( \left\|    {  \ddot{\mathbf p}}  + g \mathbf e_3 - \sum_{n = 1}^{N}  F_n \bm \rho_n /m_L \right\|, 0 \right) .
\end{align}
Denoted by $\mathcal M_{cd}$ the right hand of Eq.~\ref{equ:payload_dynamical_constraint}, and the gradients of $\mathcal M_{cd} $ w.r.t ${\mathbf c_m, T_m}$ are computed as 
\begin{subequations}
    \label{equ:thrust partial derivative}
    \begin{align}
        \frac{\partial \mathcal M_{cd}}{\partial \mathbf c_{\mathbf p}}  =&  \left( \frac{\partial \mathcal M_{cd}}{\partial {\ddot{\mathbf p}}} \right)^T  \otimes \ddot{\bm \beta},  \\ 
        \frac{\partial \mathcal M_{cd}}{\partial \mathbf c_{m, \bm{\rho}_n}} = & \left[ \left( \frac{\partial \mathcal M_{cd}}{\partial {\bm \rho}_n} \right)^T \frac{\partial {\bm \rho}_{n}}{\partial \bm \Phi_n} \right] \otimes \bm \beta,  \\
        \frac{\partial \mathcal M_{cd}}{\partial \mathbf c_{m,  F_n}} =& \frac{\partial \mathcal M_{cd}}{\partial F_n}   \bm \beta, \\ 
        \frac{\partial \mathcal M_{cd}}{\partial T_m} =& \sum_{n =1}^{N}\left[ \left( \frac{\partial \mathcal M_{cd}}{\partial \bm \rho_n} \right)^T  \frac{\partial \bm \rho_{n}}{\partial \bm \Phi_n} \dot{\bm \Phi}_n + \frac{\partial \mathcal M_{cd}}{\partial F_n} \dot{F}_n \right] \notag \\
        & + \left( \frac{\partial  \mathcal M_{cd}}{\partial \ddot{\mathbf p}} \right)^T  \dddot{\mathbf p}.   
    \end{align}
\end{subequations}

\subsection{Cable's Vectorial Constraints}
Considering an acceleration $\mathbf a$ and a set $\digamma = \{F_1 \bm \rho_1, \cdots, F_N \bm \rho_N\}$ that contains all cables's force vectors satisfying Eq.~\ref{equ:payload_model}, we find that $\mathbf a$ and a new set $\digamma' = \{R(\phi) F_1 \bm \rho_1, \cdots, R(\phi) F_N \bm \rho_N\}$ obtained by rotating all the vectors in $\digamma$ around the vector $\mathbf a + g \mathbf e_3$ by an arbitrary angle $\phi$, still satisfy Eq.~\ref{equ:payload_model}, where $R(\phi) \in SO(3)$ is the rotation matrix. Besides, from Eq.~\ref{equ:payload_model}, we also conclude that the set $\digamma'' = \{F_{\sigma(1)} \bm \rho_{\sigma(1)}, \cdots, F_{\sigma(N)} \bm \rho_{\sigma(N)}\}$ obtained by commutating the force vectors among the cables,  doesn't vary the payload's dynamics, where $\sigma: \mathcal Z^{+} \to \mathcal Z^{+}$ is the index commutation operator. Infinite feasible sets like $\digamma'$ and $\digamma''$ increase the possibility of the optimization variables getting trapped in local optimums, which gives rise to undesirable trajectories. 

To alleviate the local optima and speed up the convergence of the optimization, we separate the vectorial range of the cables and limit the optimization of each force vector within its own vectorial range. Concretely, for each $ n \in [1, \cdots, N] $, we construct the following vectorial constraints 
\begin{subequations}
    \label{equ:cable_direcional_constraint}
\begin{align}
    0 &\leq  \theta_n \leq \theta_{max}, \\
    - \frac{(2n - 3)\pi}{N}  &\leq  \phi_n \leq \frac{(2n -1) \pi }{N},
\end{align}
\end{subequations}
where $ \theta_{max} $ is the maximum allowable pitch angle of the cable. Besides, we also enforce the constraint on each cable's force $F_n$
\begin{equation}
        \label{equ:cable_force_constraint}
        F_{min} \leq  F_n \leq F_{max}, 
\end{equation}
where $F_{min}, F_{max}$ are respectively the minimum and maximum allowable forces of the cable.
This constraint prevents the cable from being slack and providing an excessive force compared to other cables.

\section{Spatial-temporal Trajectory Optimization}
\label{sec:trajectory optimization}

\subsection{System-level Path Planning}
\label{sec:safe path finding}
\begin{figure}[t]
    \begin{center}
         \includegraphics[width=0.95\columnwidth]{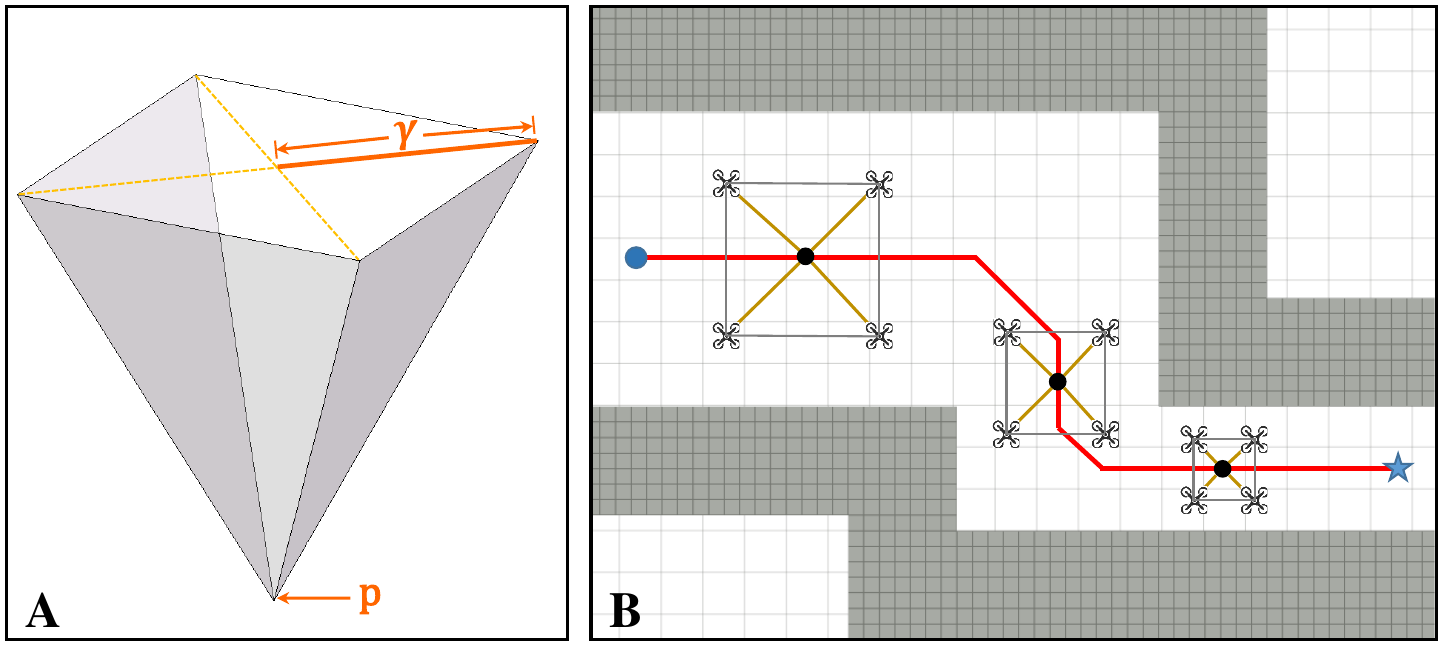}
    \end{center}
    \caption{Illustration of the system-level path planning method proposed in Sec.~\ref{sec:safe path finding}. (A) The simplified configuration of the MARTS. (B) The simplified configuration is scaled to fit different widths of the corridor.}\label{fig:30}
\end{figure}

To provide a reasonable initial path for trajectory optimization, we design a system-level global path planning method. The MARTS is simplified to a mobile and scalable regular pyramid and the length of its lateral edge is equal to the length of the cable. As shown in Fig.~\ref{fig:30}, $\gamma$ is defined as the scalable variable used to regulate the size of the regular polygon base. Let $\mathcal C_{rp}$ be the configuration space of this regular pyramid, thus given the payload's position $\mathbf p$ and $\gamma$, any configuration in $\mathcal C_{rp}$ can be uniquely represented by an ordered pair $(\mathbf p, \gamma)$. We discretize $\mathcal C_{rp}$ and use $A^*$ algorithm to obtain a collision-free path. A collision-free configuration on this path signifies that the entire regular pyramid with $\gamma$ at point $\mathbf p$ doesn't collide with any environmental obstacle. We use a solid, convex, and even conservative regular pyramid instead of a hollow, non-convex compact geometry just enveloping the entire MARTS to check the collision such that a safe corridor is not penetrated by any obstacle can be opened up by uniting the consecutive convex regular pyramids for trajectory deformation in the back-end optimization.   

\subsection{Trajectory Optimization Problem Formulation}
\label{subsec:traj opt}

The trajectory optimization problem of SAAT can be formulated as follows
\begin{subequations}
\label{eq:traj opt problem}
  \begin{align}
    \min_{\mathbf w, \mathbf T}  \label{eq:cost function}~\mathcal J_E = &\int_{0}^{T} {\norm{\mathbf p^{(s)}(t)}^2} \df{t} + \lambda_Z \sum_{n = 1}^{N} \int_{0}^{T} {\norm{\bm \xi_n^{(s)}(t)}^2} \df{t} \notag \\  &+ \lambda_T T_{\Sigma},                                                      \\
    s.t.~          & \label{eq:para map}  (\mathbf c, \mathbf T)= \mathcal L(\mathbf w, \mathbf T),  \\
                   & \label{eq:Ti>0} T_m > 0, \ \forall m \in [1,\cdots, M], \\
                   & \label{eq:init pos}~\mathbf Z^{[s-1]}(0)=\bar{\mathbf{Z}}_{0},                                                        \\
                   & \label{eq:final pos}~\mathbf Z^{[s-1]}(T_{\Sigma})= \bar{\mathbf{Z}}_{T_{\Sigma}},                                                        \\
                   & \label{eq:inequality constraint}~\mathcal G(\mathbf Z^{[s-1]}(t)) \preceq \mathbf{0}, \forall t \in [0,T_{\Sigma}],\\
                   & \label{eq:equality constraint}~\mathcal H(\mathbf Z^{[s-1]}(t)) = \mathbf{0}, \forall t \in [0,T_{\Sigma}],
  \end{align}
\end{subequations}
where $T_{\Sigma} = \sum_{m=1}^M T_m$. Cost function (\ref{eq:cost function}) compromises the smoothness and agility of the trajectory $\mathbf Z(t)$, $\lambda_Z$ regulates the smoothness of the aerial robot's motion related to the load, and $\lambda_T$ is the time regularization parameter. Eq.~\ref{eq:para map} restricts that the parameters $\{ \mathbf c, \mathbf T \}$ representing $\mathbf Z(t)$ can only be mapped by sparse optimization variables $\mathbf w, \mathbf T$ provided by $\mathcal C_{\mathbf{MINCO}}$. Eq.~\ref{eq:Ti>0} guarantees the duration of each piece is positive. Eq.~\ref{eq:init pos} and Eq.~\ref{eq:final pos} set the trajectory's boundary conditions. Eq.~\ref{eq:inequality constraint} denotes the inequality constraints continuously imposed along the trajectory, and Eq.~\ref{eq:equality constraint} are additional equality constraints designed to ensure that the flat-output trajectory $\mathbf Z(t)$ complies with the payload's dynamics.

% imposing the initial state $\mathbf z^{[s-1]}(0) $ and final state $ \mathbf z^{[s-1]}(T_{\Sigma}) $  on $\bar{\mathbf{z}}_{0} = \{\mathbf{z}_0,...,\mathbf{z}_0^{(s-1)} \}$ and $\bar{\mathbf{z}}_{0} = \{\mathbf{z}_{T_{\Sigma}},...,\mathbf{z}_{T_{\Sigma}}^{(s-1)} \}$ respectively

Then, we derive the formulations of $\mathcal J_E$ gradient w.r.t $\mathbf c_m, T_m$ as 
\begin{subequations}
    \begin{align}
	\label{equ:Je_c}
	\frac{\partial \mathcal{J}_E}{\partial \mathbf c_{m, \mathbf p}} = & 2\left( \int_0^{T_m}\bm \beta^{(s)}(t) \bm \beta^{(s)}(t)^Tdt \right) \mathbf c_{m,\mathbf p}, \\
    \frac{\partial \mathcal{J}_E}{\partial \mathbf c_{m, \bm \xi_n}} = &2 \lambda_Z  \left( \int_0^{T_m}\bm \beta^{(s)}(t)\bm \beta^{(s)}(t)^Tdt \right) \mathbf c_{m, \bm \xi_n}, \\ 
	\label{equ:Je_T}
	\frac{\partial \mathcal{J}_E}{\partial T_m} = & \left\|{\mathbf p^{(s)}(T_m)} \right\|^2 + \lambda_Z \sum_{n = 1}^{N}  \left\|{\bm \xi_n^{(s)}(T_m)}\right\|^2 \notag  \\ 
    & + \lambda_T.
\end{align}
\end{subequations}

% \begin{figure}[t]
% 	\begin{center}
% 		\includegraphics[width=0.93\columnwidth]{figures/Fig8.pdf}
% 	\end{center}
%     \vspace{-0.5cm}
% 	\caption{
% 		\label{fig:front_end}
% 		Illustration of the occlusion-aware multi-goal path-finding method. 
% 	}
% \end{figure}

\subsection{Constraints Elimination}
\label{subsec:Constraints Elimination}
The infinite number of constraints are introduced since the inequality (Eq.~\ref{eq:inequality constraint}) and the equality (Eq.~\ref{eq:equality constraint}) are enforced over the trajectory's entire duration $T_{\Sigma}$, leading to difficulties in solving the optimization problem (Eq.~\ref{eq:traj opt problem}). Therefore, to simplify this optimization problem, we design approaches to eliminate these infinite number of constraints as well as the temporal constraints (Eq.~\ref{eq:Ti>0}) as follows.

\subsubsection{Vectorial Constraint Elimination}
\label{subsubsec:Topological Constraint Elimination}
The vectorial constraints (Eq.~\ref{equ:cable_direcional_constraint} - Eq.~\ref{equ:cable_force_constraint}) enforced on the $n^{th}$ force vector are constructed directly on $\bm \xi_n$, so that for each $m' \in [1, \cdots, M-1]$, we can eliminate these constraints enforced at the ${m'}^{th}$ junction $\bm \xi_n^{m'} \in \mathbf w_{m'}$ between the ${m'}^{th}$ piece and the $(m'+1)^{th}$ piece of trajectory via designing a smooth diffeomorphism $\mathcal S: \mathbb R \to \mathbb R$. That is, let us introduce an auxiliary variable $\bm{\eta}_n^{m'} \in \mathbb R^3$, and for each element $\xi \in \bm{\xi}_n^{m'}$, we select the same element $\eta \in \bm \eta_n^{m'}$ to map $\xi$ as   
\begin{align}
    \label{equ:diffeomorphism}
    \xi = \mathcal{S}(\eta) = \frac{\xi_{min} + \xi_{max}}{2} + \frac{(\xi_{max} - \xi_{min})}{ \pi} \arctan{\eta}, 
\end{align}
where $\xi_{min}, \xi_{max}$ are respectively the lower and upper bounds of $\xi$. $\mathcal S$ maps $\mathbb R$ to interval $[\xi_{min}, \xi_{max}]$ such that arbitrary optimization of $ \eta$ on $\mathbb R$ alway guarantees $\xi \in [\xi_{min}, \xi_{max}]$.  

Then the gradient of any penalty function $\mathcal J$ w.r.t $\eta$ can be computed as
\begin{align}
    \frac{\partial \mathcal J }{\partial \eta}  = \frac{\xi_{max} - \xi_{min}}{\pi (\eta^{2} + 1)} \frac{\partial \mathcal J }{\partial \xi}.
\end{align}
% where 
% \begin{align}
%     \frac{\partial \mathcal S(\eta)}{\partial \eta} =  \frac{\xi_{max} - \xi_{min}}{\pi (\eta^{2} + 1)}. 
% \end{align}
\begin{figure}[t]
    \begin{center}
         \includegraphics[width=0.95\columnwidth]{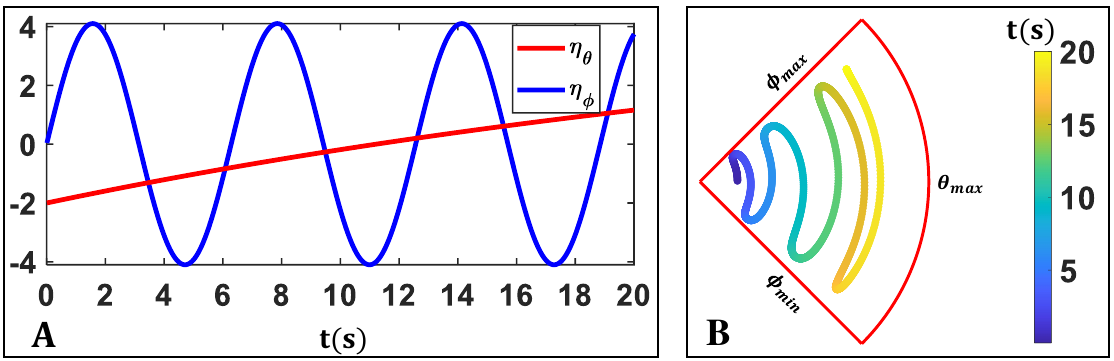}
    \end{center}
    \caption{Illustration of the diffeomorphism defined in Eq.~\ref{equ:diffeomorphism}. (A) The selected trajectories for the auxiliary variables $\eta_\theta\in \mathbb R, \eta_\phi \in \mathbb R$ to solve the trajectories of $\theta, \phi$ by Eq.~\ref{equ:diffeomorphism}. (B) The trajectory of $\bm \rho_n$ solved by Eq.~\ref{equ:rhon} using the trajectories of $\theta, \phi$ can always be compressed into a bounded red fan-shaped region.}\label{fig:OE_metric}
\end{figure}

\subsubsection{Temporal Constraint Elimination}
\label{subsubsec:Temporal Constraints Elimination}
In order to eliminate the constraint Eq.~\ref{eq:Ti>0} on $T_m$, we introduce $ \mathbf \tau = [\tau_1, \cdots, \tau_M]^T \in \mathbb R^M$ as an auxiliary temporal vector and define the following diffeomorphism to map $T_m$ from $\tau_m$
\begin{align}
    T_{m} & = e^{\tau_m},
\end{align}
such that optimizing $\tau_m$ over $\mathbb R$ always guarantees $T_m > 0$.

\subsubsection{Infinite Continuous-time Constraint Elimination}
\label{subsubsec:continuous Constraints Elimination}
Inspired by work \cite{jennings1990computational}, to transform infinite inequality constraints to finite constraints, we adopt the constraint transcription by introducing a  new integral-type constraint as
\begin{align}
      &\mathcal{J}_{\mathcal{G}}^{\int} =\int_{0}^{T_{\Sigma}}  \mathcal{J}_{\mathcal G} dt \leq 0.
\end{align}
Since the analytic value of $\mathcal{J}_{\mathcal{G}}^{\int} $ is always difficult to evaluate, we turn to approximating it by quadrature using uniformly sampled penalty functions along the trajectory as
\begin{equation}
    \label{equ:quadrature1}
    \mathcal{J}_{\mathcal{G}}^{\int} \approx  \mathcal{J}_{\mathcal{G}}^{\Sigma} = \sum_{m = 1}^{M} \mathcal{J}_{\mathcal G, m}^{\Sigma},
\end{equation}
where $\mathcal{J}_{\mathcal G, m}^{\Sigma}$ is the approximate quadrature for the $m^{th}$ piece can be calculated as 
\begin{equation}
    \label{equ:quadrature2}
    \mathcal{J}_{\mathcal G, m}^{\Sigma}  = \frac{T_m}{\kappa} \sum_{k=0}^{\kappa} \bar{\omega}_k  \mathcal{J}_{\mathcal{G}} \left(\mathbf Z|_m^{[s-1]}(t_k)\right).
\end{equation}
${\kappa}$ is the sampled number on each piece, $\left(\bar{\omega}_0, \bar{\omega}_1, \ldots, \bar{\omega}_{\kappa_i-1}, \bar{\omega}_{\kappa_i}\right)=(1 / 2,1, \cdots, 1,1 / 2)$ are the coefficients of the sampled penalty functions for the quadrature following the trapezoidal rule \cite{Press2007numerical} and $t_k=\frac{k}{\kappa} T_m$. Then the gradients of $\mathcal{J}_{\mathcal G,m}^{\Sigma}$ w.r.t. $\mathbf{c}_m$ and $T_m$ can be easily derived as
\begin{subequations}
    \begin{align}
        \label{equ:quadrature_gradient1}
        \frac{\partial \mathcal{J}_{\mathcal G,m}^{\Sigma}}{\partial \mathbf{c}_m}&= \frac{T_m}{\kappa} \sum_{k=0}^{\kappa}  \bar{\omega}_k  \frac{\partial \mathcal{J}_{\mathcal G}\left(\mathbf Z|_m^{[s-1]}(t_k)\right)}{\partial \mathbf{c}_m} , \\
        \label{equ:quadrature_gradient2}
        \frac{\partial \mathcal{J}_{\mathcal G,m}^{\Sigma}}{\partial T_m}&=\frac{\mathcal{J}_{\mathcal G,m}^{\Sigma}}{T_m}+ \frac{T_m}{\kappa^2}  \sum_{k=0}^{\kappa} k \bar{\omega}_k   \frac{\partial \mathcal{J}_{\mathcal G}\left(\mathbf Z|_m^{[s-1]}(t_k)\right)}{\partial t_k} .
    \end{align}
\end{subequations}

Since the DOF of optimization variable $ \mathbf  Z$ related to coupling dynamic constraint (Eq.~\ref{equ:payload_dynamical_constraint}) is $3 + 3 \times N$, which is much larger than $3$, i.e., the number of this constraint, we can make the trajectory approximately satisfy Eq.~\ref{equ:payload_dynamical_constraint} by constructing a finite constraint $\mathcal{J}_{\mathcal{H}}^{\int} \leq 0$ similar to Eq.~\ref{equ:quadrature1} - Eq.~\ref{equ:quadrature2} and approximately calculate $\mathcal{J}_{\mathcal H}^{\Sigma}, \mathcal{J}_{\mathcal H, m}^{\Sigma}$ similar to Eq.~\ref{equ:quadrature1} - Eq.~\ref{equ:quadrature2}. Besides, the gradients of $\mathcal{J}_{\mathcal H, m}^{\Sigma}$ w.r.t $\mathbf{c}_m$ and $T_m$ are similar to Eq.~\ref{equ:quadrature_gradient1} - Eq.~\ref{equ:quadrature_gradient2}, which are abandoned here. 
% \begin{subequations}
%     \begin{align}
%         &\mathcal{J}_{\mathcal{H}}^{\int} \approx  \mathcal{J}_{\mathcal{H}}^{\Sigma} = \sum_{m = 1}^{M} \mathcal{J}_{\mathcal H, m}^{\Sigma},   \\ 
%         &\mathcal{J}_{\mathcal H, m}^{\Sigma}  = \frac{T_m}{\kappa} \sum_{k=0}^{\kappa} \bar{\omega}_k  \mathcal{J}_{\mathcal{H}} \left(\mathbf z|_m^{[s-1]}(t_k)\right).
%     \end{align}
% \end{subequations}

\subsection{Unconstrained NLP formation}
\label{subsec:Unconstrained NLP formation}
After the elimination of the vectorial constraints in Sec.~\ref{subsubsec:Topological Constraint Elimination}, the continuous-time penalty functions for SAAT in complex environments can be summarized as
\begin{subequations}
\begin{align}
    \mathcal{J}_{\mathcal G} = & \sum_{n = 0}^{N} \lambda_{oa} \mathcal J_{oa}^{n} + \sum_{\upsilon \in \Upsilon} \sum_{n = 1}^{N} \lambda_{\mathcal \upsilon} \mathcal J_{\upsilon}^{n} + \lambda_{ra} \mathcal J_{ra},\\
    \mathcal{J}_{\mathcal H} =& \lambda_{d} \mathcal J_{d},
\end{align}
\end{subequations}
where the set of subscript is defined as $\Upsilon = \{\mathbf v, \mathbf f, \vartheta, \bm \omega\}$. $\lambda_{oa},  \lambda_{ra}, \lambda_{\mathbf v},  \lambda_{\mathbf f}, \lambda_{\vartheta}$, and $\lambda_{\bm \omega}$ are preset weights for the penalty functions. Besides,  the actual sparse trajectory parameters $\{\mathbf w, \mathbf T\}$ will be substituted by the auxiliary sparse parameters $\{\bm \varpi, \bm \tau\}$, where  $\bm \varpi = (\bm \varpi_1, \cdots, \bm \varpi_{M-1})^T \in \mathbb{R}^{(4N + 3)\times (M - 1)}$ and for any $m' \in [1, \cdots, M-1]$,  $\bm \varpi_{m'} \in \mathbb{R}^{4N + 3} $ can be denoted by $\bm \varpi_{m'} = ({\mathbf{p}^{m'}}^T, {\bm{\eta}_1^{m'}}^T, \psi_1^{m^{\prime}},  \cdots , {\bm{\eta}_{N}^{m'}}^T, \psi_N^{m'})^T$.

Then, based on the constraint elimination introduced in Sec. \ref{subsubsec:continuous Constraints Elimination}, we convert original optimization problem defined in Eq.~\ref{eq:traj opt problem} to an unconstrained optimization problem
\begin{equation}
    \label{equ:unconstainted problem}
        \begin{aligned}
            \min_{
                  \begin{scriptsize}
                    \bm {\varpi},\bm {\tau}
                  \end{scriptsize}
                 }  
              ~\mathcal J_E + \sum_{m = 1}^{M} \left(\mathcal{J}_{\mathcal G, m}^{\Sigma} + \mathcal{J}_{\mathcal H, m}^{\Sigma} \right).
        \end{aligned}
\end{equation}

This problem can be efficiently solved by fast spatio-temporal deformation provided by $\mathcal C_{\mathbf{MINCO}}$. Each deformation involves the following two procedures as shown in Fig.~\ref{fig:system_architecture}. In the forward trajectory generation, $\{\mathbf w,\mathbf T\}$ is firstly mapped from $\{\bm \varpi,\bm \tau\}$, then $\mathcal C_{\mathbf{MINCO}}$ use the banded matrix PLU decomposition to solve Eq.~\ref{equ:MCB}, acquiring $\{\mathbf c,\mathbf T\}$ from $\{\mathbf w,\mathbf T\}$. Besides, in the backward gradient propagation, through ingenious reuse of the solved PLU matrixes, $\mathcal C_{\mathbf{MINCO}}$ avoids the inversion of a $2sM$-dimensional matrix such that the gradients $\{\frac{\partial \mathcal J}{\partial \mathbf c}, \frac{\partial \mathcal J}{\partial \mathbf T}\}$ are firstly propagated to $\{\frac{\partial \mathcal J}{\partial \mathbf w}, \frac{\partial \mathcal J}{\partial \mathbf T}\}$ and further propagated to $\{\frac{\partial \mathcal J}{\partial \bm \varpi}, \frac{\partial \mathcal J}{\partial \bm \tau}\}$. Both the two processes rely just on linear spatio-temporal computational complexity. Thus, fast spatio-temporal deformation of trajectory will be performed via the updation of $\{\bm \varpi,\bm \tau\}$.

\section{Control Scheme for Agile Transportation}
In this work, we propose a robust and distributed control scheme to achieve agile transportation, the architecture of which is shown at the bottom of Fig.~\ref{fig:system_architecture}. It should be pointed out that in this section, we add $n$ to the superscript or subscript of symbols to specify that they belong to the $n^{th}$ aerial robot, but for simplicity, we omit $n$ in Fig.~\ref{fig:system_architecture}. This scheme gets rid of the reliance on the state measurement for both payload and cable as in works \cite{li2021cooperative, geng2020cooperative, li2023nonlinear, wahba2024efficient}, as well as the closed-loop control for payload as in works \cite{lee2013geometric, lee2017geometric, li2021cooperative, geng2020cooperative, li2023nonlinear, wahba2024efficient}. Instead, it pursues closed-loop control for each aerial robot. That's to say the position and velocity errors of the payload would not be considered in the aerial robot's control law such that the system oscillation induced by the payload's control law and time-delay induced by control signal transmission can be avoided fundamentally. INDI is used to estimate the direction and magnitude of the tension in each cable such that we don't need to deploy additional tension sensors or attitude sensors to measure these states, and the unknown payload's mass can be online estimated. Both the actual estimation for the cable's force vector and planned trajectory are used to calculate the flatness maps and further get the desired states. Since the INDI always compensates the actual force exerted by the cable, this control scheme is robust against the deviations from the planned trajectory, which are introduced by the payload's swing around the cables' attachment point induced by the inevitable attachment error and the estimation error of the payload's mass. The details about the outer-loop trajectory tracking controller and the inner-loop geometric attitude controller contained in this scheme are as follows.

\subsection{Outer Loop Controller}
\label{subsec:Outer Loop Controller}
Before we derive the control law, the actual mass-normalized thrust $f_n$ and control torque $\bm \tau_n$ of the $n^{th}$ aerial robot should be firstly estimated by the mixing matrix using the RPMs of motors, i.e., $r_{n,1},\cdots, r_{n,4}$. 

For the outer loop, the external forces that each aerial robot suffers from, such as the actual tension exerted by the cable and the aerodynamic drags are considered as a whole. The feedforward trajectory tracking control law for the $n^{th}$ aerial robot is calculated as 
\begin{align}
    \mathbf a_{n,c}  = K_p(\mathbf p_{n,d} - \mathbf p_n) + K_v(\mathbf v_{n,d} - \mathbf v_n) + \mathbf a_{n,d}.
\end{align}
Based on the principle of INDI, the mass-normalized thrust vector command $ \mathbf f_{n,c}$ that provides $\mathbf a_{n,c}$ can be computed using the following incremental relation
\begin{align}
    \mathbf f_{n,c} = \mathcal{FL}( f_{n} \mathbf z_{\mathcal B}^{n}) + \mathbf a_{n,c} - \mathcal{FL}(\mathbf a_n),
\end{align}
where $\mathcal{FL}()$ is a low-pass filter. Then the mass-normalized thrust command $ f_{n,c} $ and its unit vector direction $ \mathbf z_{\mathcal B, c}^{n} $ are obtained as 
\begin{equation}
    f_{n,c} = \|\mathbf f_{n,c}\|_2, \quad \mathbf z_{\mathcal B, c}^{n} = \mathcal N (\mathbf f_{n,c}).
\end{equation}
Besides, the tension $-\widetilde{F_n \bm \rho_n}$ exerted on the $n^{th}$ aerial robot by the $n^{th}$ cable is estimated by rewriting Eq.~\ref{eq:acceleration} as follows
\begin{equation}
-\widetilde{F_n \bm \rho_n} =  -m_n (\mathcal{FL} ({\mathbf a}_n ) - g \mathbf e_3 +  \mathcal{FL} (f_n \mathbf z_{\mathcal B}^{n})).
\end{equation}
Then, the $n^{th}$ cable's tension ${F_n}$ and the vector direction $\bm \rho_{n}$ can be estimated as 
\begin{equation}
    \widetilde{F_n} = \| \widetilde{F_n \bm \rho_n} \|_2, \quad \widetilde{ \bm \rho_{n}} = \mathcal N (\widetilde{F_n\bm \rho_n}).
\end{equation}

It should be pointed out that $-\widetilde{F_n \bm \rho_n}$ contains not only the $n^{th}$ cable's tension but also the disturbances such as the wind and the aerodynamic drag that the aerial robot suffers from. 

\subsection{Inner Loop Controller}
\label{subsec:Inner Loop Controller}
For the inner loop, since it is generally not possible to tie a cable strictly to the CoM of the aerial robot, the INDI is also used to compensate for the inevitable external moment induced by the cable. Based on the desired yaw angle $\psi_{n,d}$ and the unit vector direction of the thrust, we reconstruct the desired attitude $\mathbf {q}_{n,d}$ using the rotation factorization provided by Hopf fibration 
\begin{equation}
    \mathbf {q}_{n,d} =\mathbf {q}_{\mathbf z_{\mathcal B,c}^{n}} \odot \mathbf {q}_{\psi_{n,d}}. 
\end{equation}
Then the quaternion $\mathbf {q}_{n,e}$ representing the rotation to align the desired attitude $\mathbf {q}_{n,d}$ to current attitude $\mathbf {q}_{n}$ is calculated as
\begin{equation}
    \mathbf {q}_{n,e} = \mathbf {q}_{n}^{-1} \odot \mathbf {q}_{n,d}. 
\end{equation}
$\mathbf {q}_{n,e}$ implies a rotation around an axis $\iota_n $ denoted by $\iota_n = (\mathbf q_{n,e}^x, \mathbf q_{n,e}^y, \mathbf q_{n,e}^z)^T$, thus we define a vector $ \mathbf \Theta_{n,e} $ to describe this rotation as 
\begin{equation}
   \mathbf \Theta_{n,e} = 2 \arccos(\mathbf q_{n,e}^w) \mathcal N (\iota_n). 
\end{equation}
Next, the attitude tracking control law ${\dot{\bm \omega}_{n,c}}$ is designed as 
\begin{equation}
    \dot{\bm \omega}_{n, c} = K_{\mathbf \Theta} \mathbf \Theta_{n,e} + K_{\bm \omega} (\bm \omega_{n,e}) + K_{I} \int \bm \omega_{n,e} dt +  \dot{\bm \omega}_{n,d}. 
 \end{equation}
where $\bm \omega_{n,e} $ is the error of body rate that can be calculated as
\begin{equation}
    \bm \omega_{n, e} =  \mathbf {q}_{n}^{-1} \odot  \mathbf {q}_{n,d}  \odot \bm \omega_{n,d} - \mathcal{FL}( \bm {\omega}_n). 
 \end{equation}
Note that the desired body rate $\bm \omega_{n,d}$ is calculated by Eq.~\ref{equ:omega1} - Eq.~\ref{equ:omega3} based on the states getting from the trajectory except that $F_n, \bm \rho_{n}$ used to calculate $\mathbf{z}_{\mathcal{B}}^{n}$ (Eq.~\ref{eq:zb}) and $\mathbf{z}_{\mathcal{B}}^{n}$ (Eq.~\ref{eq:dzb}) are substituted by their estimations $\widetilde{F_n}, \widetilde{ \bm \rho_{n}}$. From the primitive of INDI, the incremental expression for the moment command can be computed as
\begin{equation}
    \bm {\tau}_{n,c} = \mathcal{FL}(\bm \tau_n)  + \mathbf J_{n} (\dot{\bm \omega}_{n,c} - \dot{\bm {\omega}}_n). 
 \end{equation}
It should be pointed out that the low-pass filters for $\mathbf a_n$, $f_n \mathbf z_{\mathcal B}^{n}$, $\bm \omega_n$ and $ \bm \tau_n$ are set to the same cut-off frequency $cof$.
 
 \subsection{Estimation for Mass of Payload}
 \label{subsec:Estimation for Mass of Payload}
 When the payload is lifted, the MARTS will be stabilized at a predefined position. Then the mass of the payload can be estimated by a finite sequence of each aerial robot's estimations for the force vector of the cable attached to it as follows
 \begin{equation}
     \label{equ:mass estimation}
     \widetilde{m_L} = \sum_{n = 1}^{N} \sum_{ w = 1}^{W} \frac{\widetilde{F_n \bm \rho_n}|_{w,3}}{W}  . 
  \end{equation}
 where $\widetilde{F_n \bm \rho_n}|_{w,3}$ denotes the z component of the $w^{th}$ estimation for $F_n \bm \rho_n$ in the sequence and $W$ denotes the number of the estimations in the sequence. 
 
 \begin{figure*}[t]
    \begin{center}
         \includegraphics[width=1.88\columnwidth]{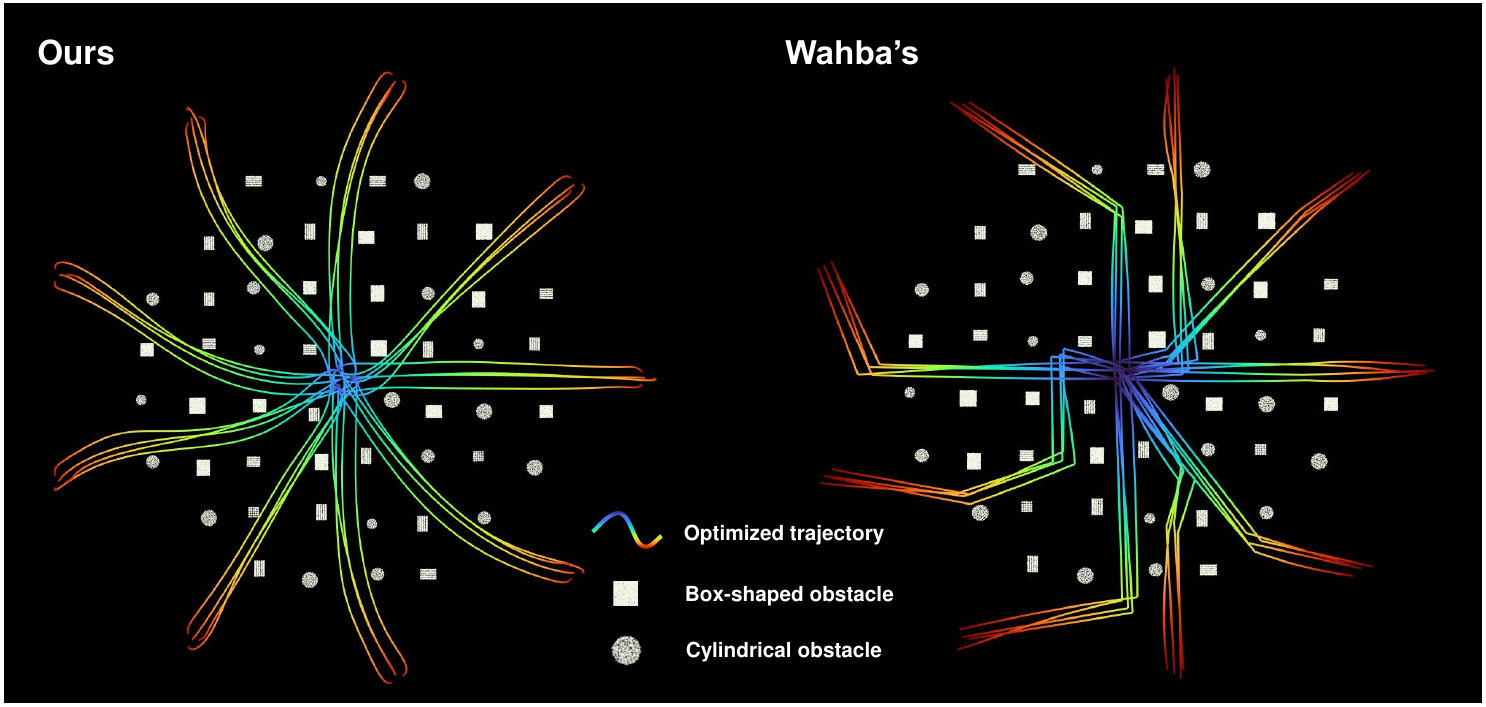}
    \end{center}
    \caption{Comparison of the partial trajectories  in the medium-density environment optimized by the two methods. }\label{fig:trajectory comparision}
\end{figure*}

 \begin{table}[]
     \caption{Parameter Setting for Simulations and Experiments}
      \label{tab:para setting}
     \centering
     \renewcommand{\arraystretch}{1.3}
     \begin{tabular}{|llllll|}
     
     \hline
 
     \multicolumn{6}{|c|}{System Parameter}    \\ \hline
 
     \multicolumn{1}{|l|}{$m_i$} &  \multicolumn{2}{c|}{320g} & \multicolumn{1}{l|}{$l$}  &  \multicolumn{2}{c|}{1.2m}     \\ \hline
 
     \multicolumn{1}{|l|}{$J_i$} &  \multicolumn{5}{c|}{$diag(4.463,4.725,5.340) \times 10^{-4}kg \cdot m^2$}  \\ \hline
 
     \multicolumn{6}{|c|}{Planning Parameter}      \\ \hline
 
     \multicolumn{1}{|l}{$d_{oa}^L$}   &  $0.2m$ & \multicolumn{1}{|l}{$d_{oa}^Q$} & $0.3m$  & \multicolumn{1}{|l}{$d_{oa}^c$}  &  $ 0.2m $  \\ \hline
 
         \multicolumn{1}{|l}{$K$}   &  7 & \multicolumn{1}{|l}{$d_s$} & 0.2m  & \multicolumn{1}{|l}{$v_{max}$}  &  $6.0m/s$  \\ \hline
 
     \multicolumn{1}{|l}{$f_{max}$} & $30 N/kg$ & \multicolumn{1}{|l}{$\vartheta_{max}$}   &  $1.05rad$ & \multicolumn{1}{|l}{$\omega_{max}$} & $2.7rad/s$ \\ \hline
 
     \multicolumn{1}{|l}{$\theta_{max}$}   & $1.0 rad$ & \multicolumn{1}{|l}{$F_{min}$} & $0.24N$  &  \multicolumn{1}{|l}{$F_{max}$} & $2.4N$   \\ \hline
     
     \multicolumn{1}{|l}{$\lambda_T$} & $2000.0$   &  \multicolumn{1}{|l}{$\lambda_z$} & $0.3$ & \multicolumn{1}{|l}{$\lambda_{oa}$} & 10000.0 \\ \hline
 
     \multicolumn{1}{|l}{$\lambda_{ra}$} & $10000.0$ & \multicolumn{1}{|l}{$\lambda_{v}$} & 1000.0  &\multicolumn{1}{|l}{$\lambda_{\tau}$}   & 1000.0   \\ \hline
     
     \multicolumn{1}{|l}{$\lambda_{\omega}$} & 1000.0  & \multicolumn{1}{|l}{$\lambda_{\vartheta}$}   &  1000.0 & \multicolumn{1}{|l}{$\lambda_{\varsigma}$} & 10000.0  \\ \hline
 
     \multicolumn{6}{|c|}{Control Parameter}    \\ \hline
 
     \multicolumn{1}{|l}{$cof$} & \multicolumn{2}{l|}{20Hz} &  \multicolumn{1}{l}{$W$} & \multicolumn{2}{l|}{1000}   \\ \hline
     
     \multicolumn{1}{|l}{$K_p$}  & \multicolumn{2}{l|}{$diag(12.0,12.0,3.0)$}  & \multicolumn{1}{l}{$K_v$}  &  \multicolumn{2}{l|}{$diag(4.0,4.0,2.0)$}   \\ \hline
     
     \multicolumn{1}{|l}{$K_{\mathbf \Theta}$}   &  \multicolumn{2}{l|}{$diag(70.0,100.0,19.0)$} &\multicolumn{1}{l}{$ K_{\bm \omega}$}    &  \multicolumn{2}{l|}{$diag(10.0,12.0,3.0)$}  \\ \hline
     
     \multicolumn{1}{|l}{$ K_I$}    &  \multicolumn{2}{l}{$diag(0.0,0.0,0.3)$}  &  \multicolumn{3}{l|}{}  \\ \hline

     \end{tabular}
 \end{table}
 
 \begin{table}[b]
    \centering
    \caption{Performance Comparison between the Two Methods}
    \label{tab:Performance Comparison between Planners}
    \renewcommand{\arraystretch}{1.25}
    \begin{tabular}{cccc}
    \hline
    Scenario                & Method  & success rate(\%) & length($m$) \\ \hline
    \multirow{2}{*}{Sparse} & Wahba's &   83.3  &  22.069   \\ \cline{2-4} 
        & Ours    &     $\bm{100}$    &   $\bm{21.422}$        \\ \hline
    \multirow{2}{*}{Medium} & Wahba's &    66.8  &  23.477    \\ \cline{2-4} 
     & Ours    &   $ \bm{100}$   &   $\bm{21.568}$     \\ \hline
    \multirow{2}{*}{Dense}  & Wahba's &     44.5   &  25.282   \\ \cline{2-4} 
     & Ours    &   $\bm{100}$   &   $\bm{22.211}$      \\ \hline
    \end{tabular}
\end{table}

\section{Benchmarks and Simulations}
In this section, benchmarks and simulations are carried out to validate our proposed trajectory planning scheme for SAAT in complex environments. First, we compare our planning scheme with a SOTA planner and analyze the
performance in terms of responsiveness and trajectory quality (Sec.~\ref{sec:benchmark}). Then, we conduct an ablation study (Sec.~\ref{sec:ablation}) to validate that the vectorial constraint (Eq.~\ref{equ:cable_direcional_constraint}) enforced to the cable is indispensable to avoid undesirable locally optimal trajectory.

The optimization and visualization of the trajectories are implemented on a laptop with an Intel Core i5-10300H CPU(2.5 GHz) and don't depend on any hardware acceleration. The code implementation of our methods is based on C++11. Important parameters used for the following benchmarks, simulations in this section, and real-world experiments in Sec.~\ref{sec:Real World Experiments} are shown in Tab.~\ref{tab:para setting}.

\subsection{Benchmark Comparisons}
\label{sec:benchmark}
In this work, a SOTA kinodynamic motion planner for the payload transportation using MARTS proposed by Wahba et al. \cite{wahba2023kinodynamic} is selected, which relies on the Flexible Collision Library \cite{pan2012fcl} for collision checking and is implemented based on the differential dynamic programming \cite{10758212} solver in Crocoddyl \cite{mastalli2020crocoddyl}, to compare the performance with our trajectory planning scheme.

% \usepackage{multirow}

The first experiment tests the quality of the optimized trajectories planned by the two methods. Three circular environments with radii of $18m$, featuring sparse, medium, and dense obstacle densities are constructed, among which the medium-density environment is shown in Fig.~\ref{fig:trajectory comparision}. We select the center of the circular environment as the starting point and uniformly choose $36$ points on a circle with a radius of $21m$ as the target points to respectively plan $36$ trajectories using the two methods. The bounds of the constraints considered in both methods are set to be the same as listed in Tab~\ref{tab:para setting}. An optimized result is successful if the planner generates a dynamically feasible trajectory. For each obstacle density, we evaluate the two methods for the MARTS consisting of three aerial robots in terms of the success rate and the average length of these $36$ trajectories.   

% Since KMP cannot adjust the trajectory's duration during the optimization, we fix it to $7s$. This is mainly because the sparse discretion of system dynamics cannot guarantee the safety of the entire system between adjacent discrete states.  

The result is shown in Tab.~\ref{tab:Performance Comparison between Planners}. As the density of obstacles increases, the success rate of Wahba's method significantly decreases, whereas our method maintains a high success rate. We find that Wahba's method, when the environment is dense, is likely to fall into a dynamically infeasible local optimum, resulting in unsuccessful trajectories. 
A comparison of the successful trajectories in the medium-density environment generated by both methods is shown in Fig.~\ref{fig:trajectory comparision}. Compared to Wahba's method, our method can generate shorter trajectories, especially in the dense environment, which relies on the agility provided by the spatio-temporal deformation of $\mathcal C_{\mathbf{MINCO}}$. Besides, there exist obvious abrupt turning points in those trajectories generated by Wahba's method, whereas our trajectories are smoother.

\begin{figure}[]
    \begin{center}
         \includegraphics[width=0.95\columnwidth]{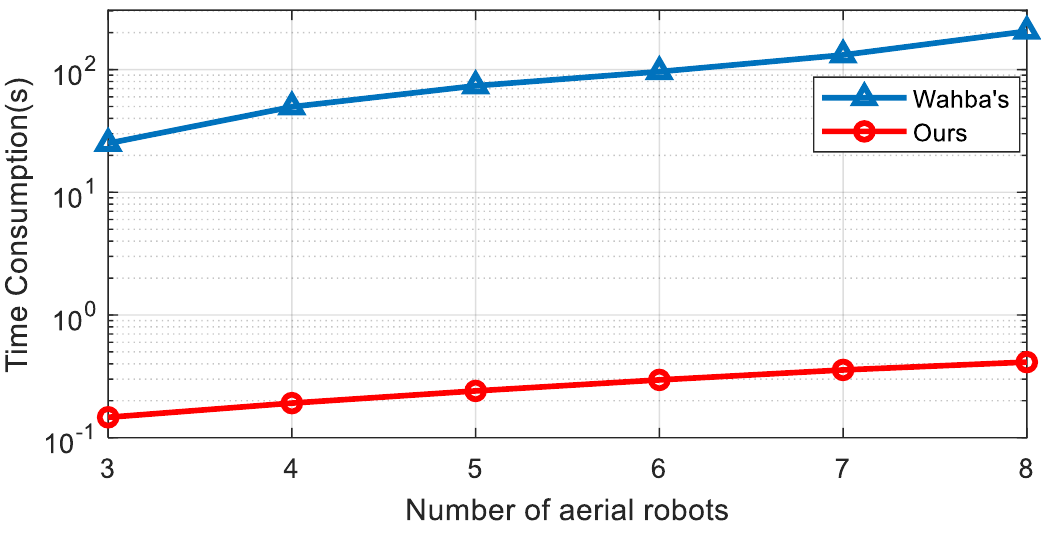}
    \end{center}
    \caption{Responsiveness comparison of the two methods.}\label{fig:Responsiveness Comparison}
\end{figure}

The second experiment tests the responsiveness of the two methods when the MARTS contains different numbers of aerial robots. This experiment is conducted in the medium-density environment. For both planning methods, the numbers of decision variables and constraints scale linearly with the number of aerial robots. Nevertheless, our method reduces the time consumption of trajectory optimization by two orders of magnitude compared to Wahba's method, and the curve of the time consumption for trajectory optimization as the number of aerial robots grows is shown in Fig.~\ref{fig:Responsiveness Comparison}. Theoretically, the time consumption of Wahba's method grows cubically, whereas our planner grows approximately linearly. The reason for the responsiveness improved by our planner is that in each iteration of optimization, as the number of aerial robots increases, the elapsed time for constraint evaluation and penalty functions calculation grows linearly and the elapsed time for trajectory generation ($\bm \varpi^{*}, \bm \tau^* \rightarrow \mathbf c, \mathbf T$ as shown in Fig.~\ref{fig:system_architecture}) hardly increases.  
 
\begin{figure}[]
    \begin{center}
         \includegraphics[width=0.95\columnwidth]{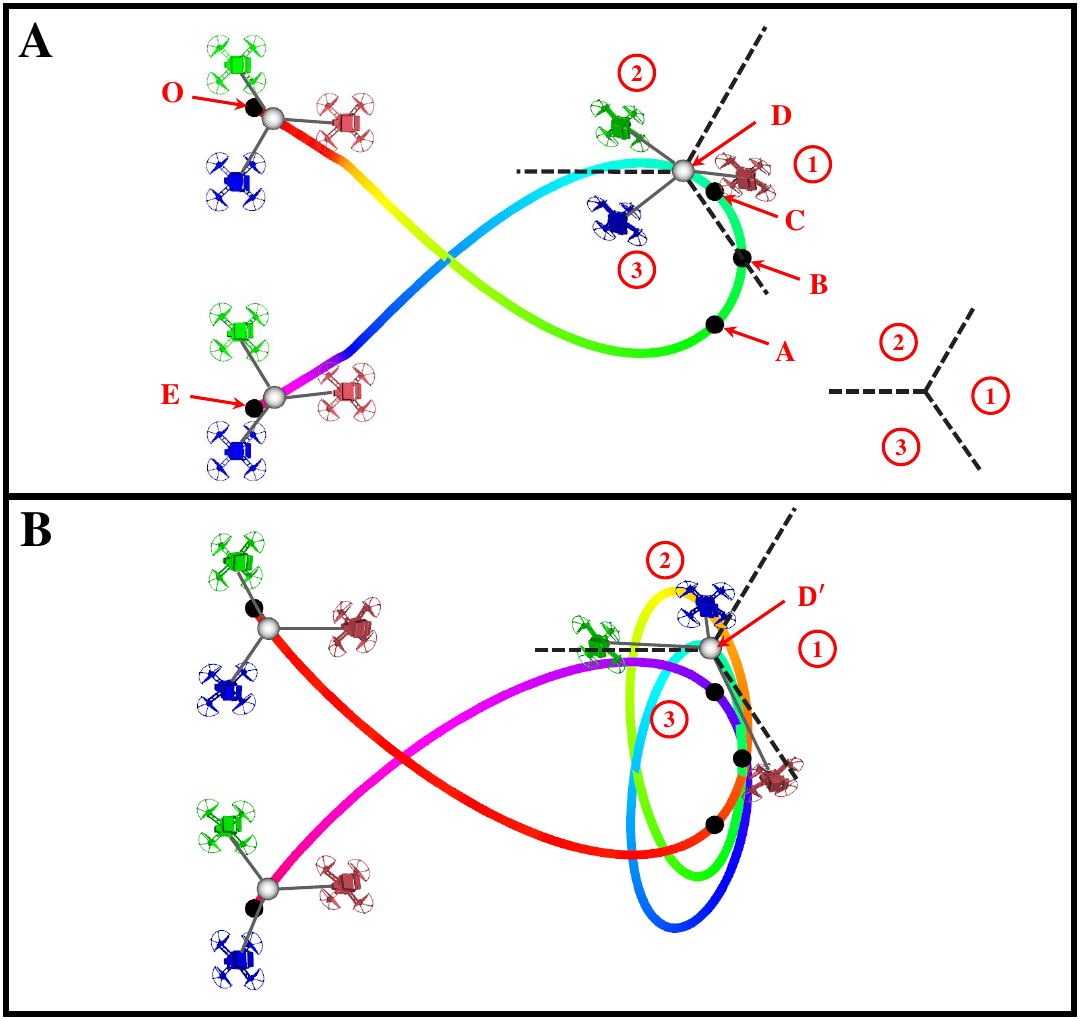}
    \end{center}
    \caption{Ablation study for the cable's vectorial constraints defined in Eq.~\ref{equ:cable_direcional_constraint}. The trajectories optimized with and without these constraints are shown in (A) and (B) respectively.}\label{fig:ablation1}
\end{figure}

\subsection{Ablation Study}
\label{sec:ablation}
The cable's vectorial constraints defined in Eq.~\ref{equ:cable_direcional_constraint} separately restrict the aerial robots as well as their attached cables to three regions, as shown in the lower right corner of Fig.~\ref{fig:ablation1}A. In this experiment, point O and point E are set as the start and target points of the payload's trajectory respectively. A, B, and C are fixed waypoints that the payload's trajectory must pass through. The proposed planning scheme with vectorial constraints optimizes the optimal trajectory with a simple topology, as shown in Fig.~\ref{fig:ablation1}A, whereas the proposed planning scheme without vectorial constraints falls into a local optimum and generates a trajectory with a complex topology, as shown in Fig.~\ref{fig:ablation1}B. A major difference between these two trajectories is that at point $D'$ in Fig.~\ref{fig:ablation1}B, the relative arrangement of the red and blue aerial robots differs from the initial relative arrangement at the start point $O$ and the final relative arrangement at the target point $E$, in contrast to the point $D$ in Fig.~\ref{fig:ablation1}A, where the three aerial robots always maintain a relative arrangement as same as those at the start point $O$ and the target point $E$. The main reason for the occurrence of the local optimum in Fig.~\ref{fig:ablation1}B is that there are infinitely many possible combinations of cables' force vectors that allow the trajectories to satisfy the payload's dynamics (Eq.~\ref{equ:payload_model}). The planner reduces $\mathcal J_d$, namely the violation of payload's dynamics too quickly by drastically deforming the aerial robots' trajectories to an undesirable combination of cables' force vectors at the expense of the trajectories' energy $\mathcal J_E$, which leads to an equilibrium between $\mathcal J_d$ and $\mathcal J_E$. At this moment, if the planner further optimizes to reduce $\mathcal J_E$, it will inevitably lead to an increase of $\mathcal J_E$, thus inducing this local optimum. Therefore, the vectorial constraints restrict the magnitude of the trajectory deformation by reducing the number of possible combinations among the cables' force vectors, which greatly alleviates the local optimum.

\begin{figure*}[t]
    \begin{center}
        \includegraphics[width=1.88\columnwidth]{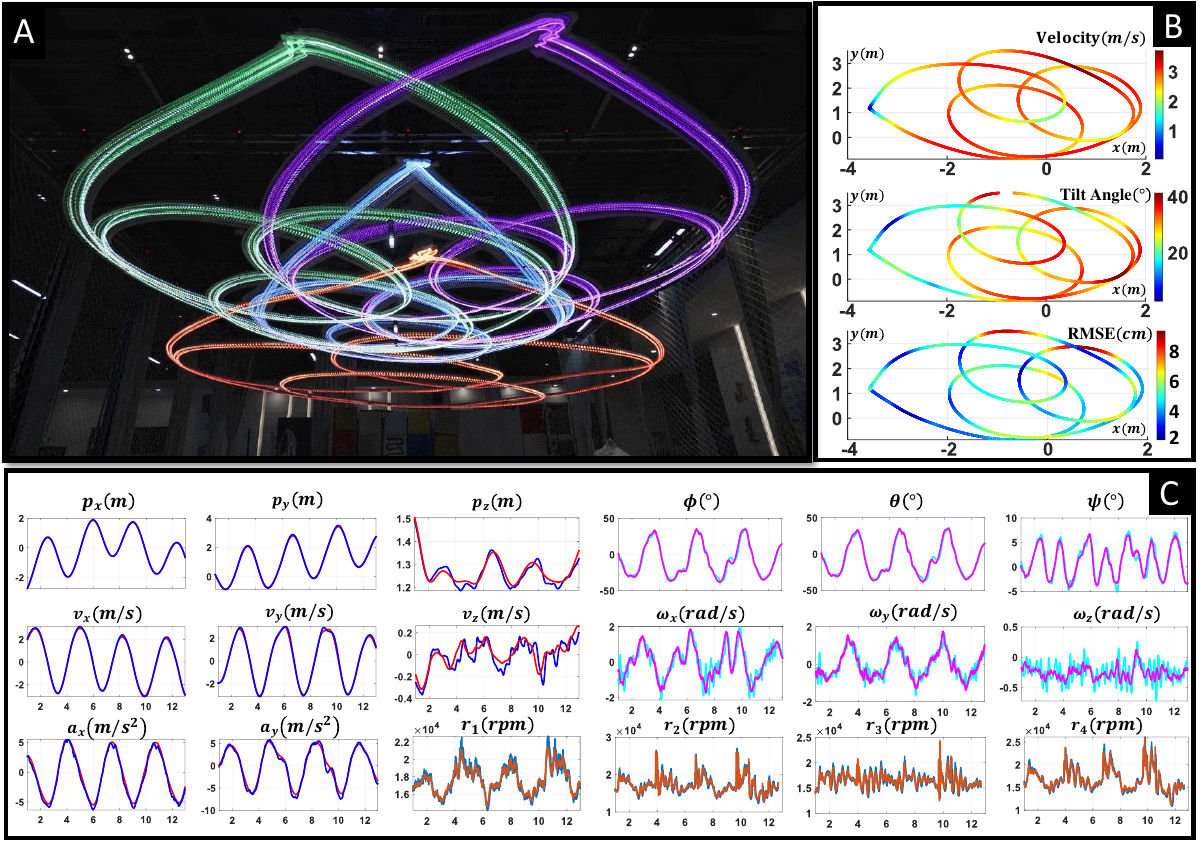}
    \end{center}
    \vspace{-0.3 cm}
    \caption{Control results for tracking agile trajectory in free space, whose agility approaches the limit that the MARTS can successfully execute. (A) Sequential snapshots of the agile transportation trajectory, in which the red curve represents the payload's trajectory and the other three curves represent the three aerial robot's trajectories. (B) The instantaneous velocity, tilt angle, and RMSE of the $2^{th}$ aerial robot are displayed on its trajectory respectively. (C) The result of trajectory tracking, including the position, velocity, acceleration, attitude, bodyrate, and rotors' RPMs. The curves using colors \textcolor[rgb]{0.85,0.33,0.1}{{$\blacksquare$}}, \textcolor[rgb]{1.0,0.0,0.0}{$\blacksquare$}, and \textcolor[rgb]{1.0,0.0,1.0}{$\blacksquare$} represent the desired states, and the curves using colors \textcolor{blue}{$\blacksquare$}, \textcolor{cyan}{$\blacksquare$}, and \textcolor[rgb]{0.0,0.45,0.74}{$\blacksquare$} represent the actual state.}
    \label{fig:control result}
\end{figure*}

\section{Real World Experiments}
\label{sec:Real World Experiments}
\subsection{System Configuration}

\begin{figure}[t]
    \begin{center}
         \includegraphics[width=0.95\columnwidth]{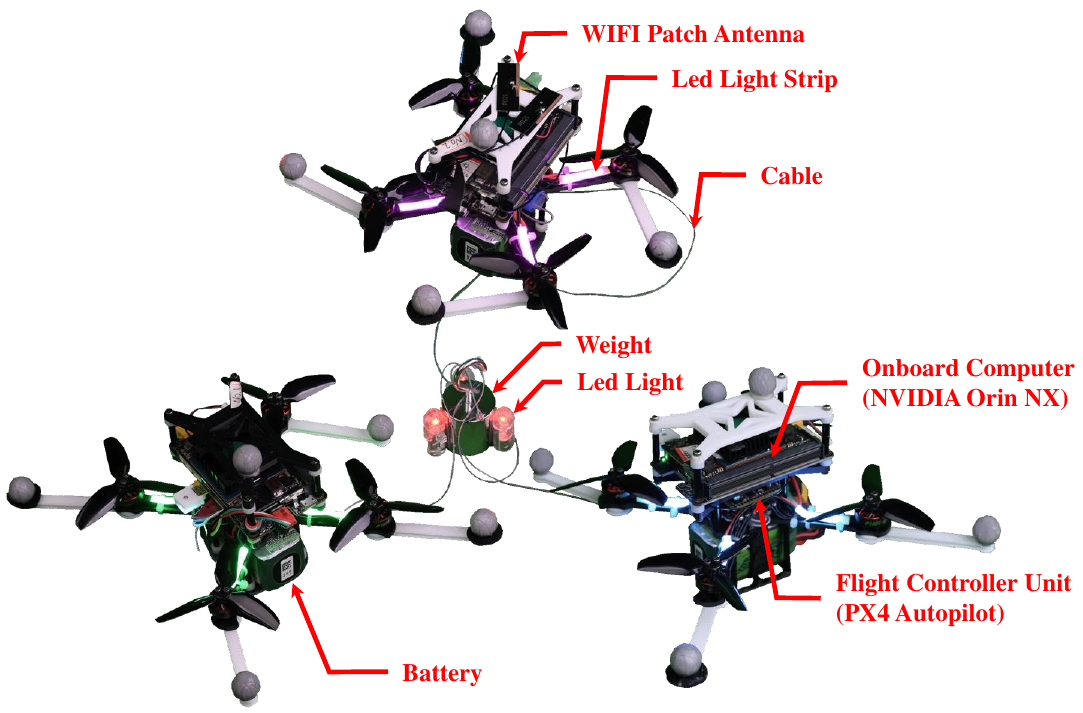}
    \end{center}
    \caption{Illustration of our MARTS consisting of three aerial robots.}\label{fig:ATS}
\end{figure}

To carry out the real experiments, we deploy a practical MARTS consisting of three aerial robots, each weighing 320g with a wheelbase of 140mm, as shown in Fig.~\ref{fig:ATS}. Each aerial robot is equipped with an NVIDIA Jetson Orin NX as the onboard computer, a Kakute H7 MINI as the flight control unit, and a WIFI module. APM firmware is flashed into the flight control unit to get the motor's speed. An EKF fusing the information from the motion capture system and the IMU is used to estimate the odometry for each aerial robot. The IMU frequency is set to 333Hz. The software modules, including odometry estimation and flight control (both the outer-loop controller and inner-loop controller) run in real-time at IMU's frequency on the NX computer. The optimized trajectory can be transmitted to each aerial robot through a broadcast network. For the experiments in complex environments, we use FAST-lio2, an excellent LiDAR-inertial odometry framework to get the point cloud data beforehand and then build the ESDF.

\subsection{Control Scheme Validation for Agile Transportation}

We validate the performance of our control scheme in the following four aspects. Firstly, the precision of the mass estimation for the payload needs to be evaluated. Secondly, we gradually increase the agility of the trajectory until any aerial robot's average motor speed approaches the limit to evaluate the tracking error of the control scheme. Thirdly, we test the robustness of the control scheme against common model uncertainties on payload existing in realistic transportation. Finally, we verify whether accurate estimation and compensation for the cable's force is necessary. We construct four test scenarios and the details are as follows.

\begin{figure}[t]
    \begin{center}
         \includegraphics[width=0.95\columnwidth]{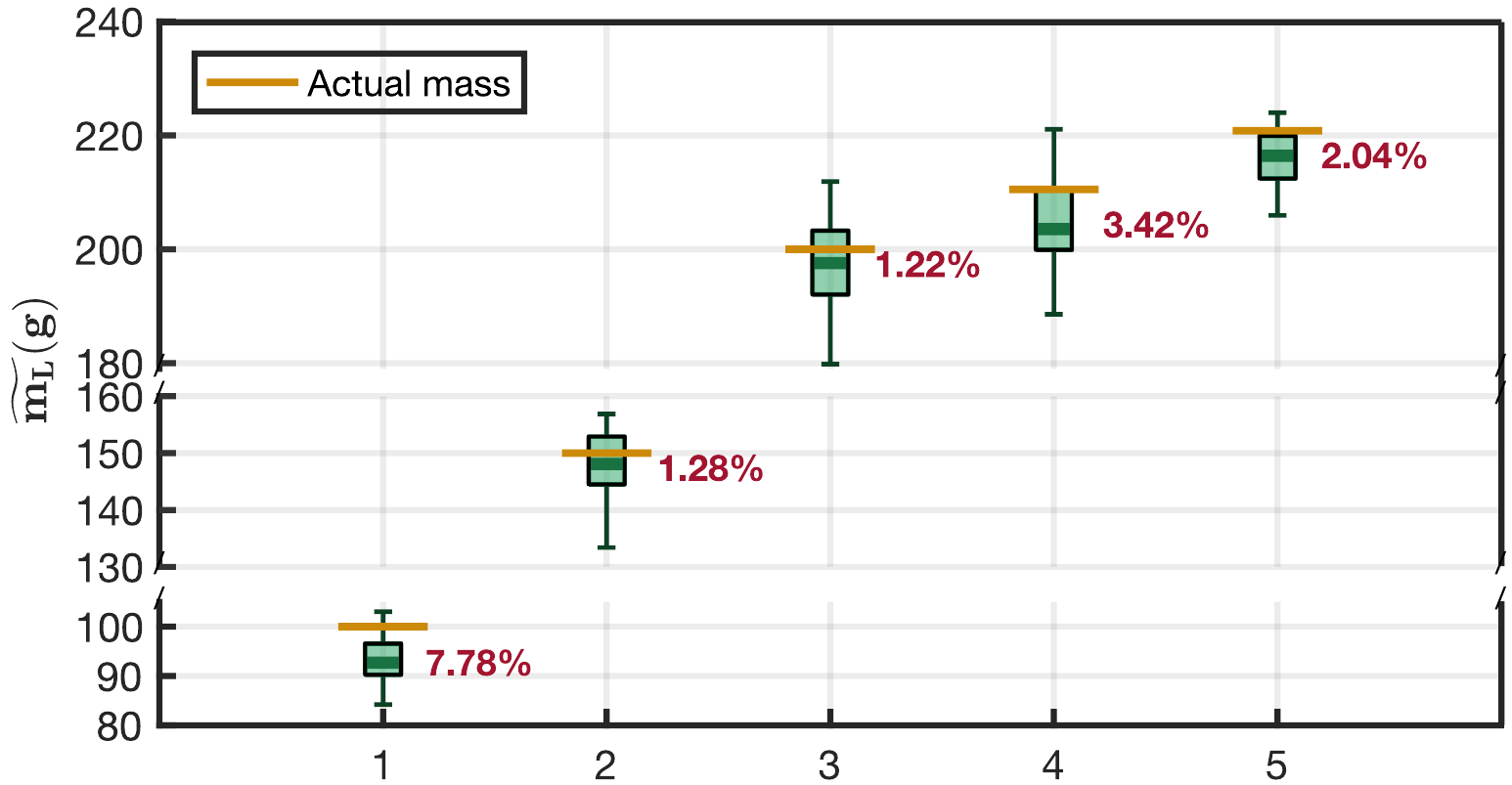}
    \end{center}
    \caption{Estimations for the masses of five different payloads.}\label{fig:payload mass}
\end{figure}

\begin{table}[b]
    \centering
    \caption{Maximum Agility Tests of the MARTS for Multiple Weights}
    \label{tab:Tracking Error of Agile Transportation}
    \renewcommand{\arraystretch}{1.25}
    \begin{tabular}{|l|c|l|l|l|l|}
    \hline
    \multicolumn{1}{|c|}{\multirow{2}{*}{$100g$}} & \begin{tabular}[c]{@{}c@{}}Max Vel($m/s$)\\ Max Acc($m/s^2$) \end{tabular} & \begin{tabular}[c]{@{}c@{}} 4.4 \\ 9.104 \end{tabular} & \begin{tabular}[c]{@{}c@{}} 4.5 \\ 9.126 \end{tabular} & \begin{tabular}[c]{@{}c@{}} 4.55 \\ 9.175 \end{tabular} & \begin{tabular}[c]{@{}c@{}} 4.6 \\ 9.184 \end{tabular} \\ \cline{2-6} 
    \multicolumn{1}{|c|}{}  & RMSE($cm$)& $\bm{4.634} $ & $\bm{4.916}$  & $\bm{5.261} $ &  $\bm{6.406}$ \\ \hline

    \multirow{2}{*}{$150g$}   & \begin{tabular}[c]{@{}c@{}}Max Vel($m/s$)\\ Max Acc($m/s^2$)\end{tabular} &  \begin{tabular}[c]{@{}c@{}} 4.0 \\ 7.278 \end{tabular}  &  \begin{tabular}[c]{@{}c@{}} 4.1 \\ 7.859 \end{tabular} & \begin{tabular}[c]{@{}c@{}} 4.15 \\ 7.942 \end{tabular}  & \begin{tabular}[c]{@{}c@{}} 4.2 \\ 8.443 \end{tabular}  \\ \cline{2-6} 
     & RMSE($cm$)    &  $\bm{4.412}$ & $\bm{4.569}$  &  $\bm{5.454}$ & $\bm{6.209} $ \\ \hline

    \multirow{2}{*}{$200g$}  & \begin{tabular}[c]{@{}c@{}}Max Vel($m/s$)\\ Max Acc($m/s^2$)\end{tabular} & \begin{tabular}[c]{@{}c@{}} 3.6 \\ 5.749 \end{tabular}   &  \begin{tabular}[c]{@{}c@{}} 3.7 \\ 6.126 \end{tabular}  &  \begin{tabular}[c]{@{}c@{}} 3.75 \\ 6.292 \end{tabular}  &  \begin{tabular}[c]{@{}c@{}} 3.8 \\ 6.458 \end{tabular}  \\ \cline{2-6} & RMSE($cm$)   &  $\bm{4.553}$ & $\bm{5.042}$  & $\bm{5.288} $ &  $\bm{5.574}$ \\ \hline
    \end{tabular}
    \end{table}

\subsubsection{Scenario 1. Estimation for Payload's Mass}
The precise mass of the payload is always not known beforehand. In this experiment, we select five payloads with different masses. The MARTS lifts the payload off the ground and then stabilizes each aerial robot to a predefined hovering position. We estimate each payload's mass 10 times using Eq.~\ref{equ:mass estimation} in Sec.~\ref{subsec:Estimation for Mass of Payload} and the results are listed in Fig.~\ref{fig:payload mass}. All the estimation errors are no more than $8\%$. In our work, we use a fixed thrust coefficient to estimate the actual thrust of the aerial robot, which leads to an estimation error of the actual thrust at different RPMs. The estimation error of the actual thrust, the measurement error of the RPM, and the payload's residual acceleration in $\mathbf z_{\mathcal I}$-axis while hovering are the main causes of errors in estimating the payload's mass. The results of this experiment show that our method can approximately estimate the mass of the payload.

\subsubsection{Scenario 2. Agile Transportation to the Limit of Aerial Robot's Thrust}
Then, we need to validate whether the proposed control scheme can hold the maximum agility of MARTS. For the MARTS, its maximum agility mainly relates to the maximum executable acceleration of the payload. Therefore, we generate a type of agile trajectory as shown in Fig.~\ref{fig:control result}A. Before entering the small circle along the trajectory, the MARTS needs to be accelerated to a necessary speed to produce a large centripetal acceleration. The increase of the circle's acceleration can be regulated by reducing the duration of the trajectory. Three weights weighing $100g, 150g$, and $200g$ respectively act as the payload. For each weight, we improve the acceleration of the trajectory as much as possible empirically until the maximum instantaneous average motor's RPM of any aerial robot in the MARTS along the trajectory approaches $24000 RPM$ (only $13600RPM$ is required for hovering without a payload). Higher RPM can cause a dramatic drop in battery voltage, leading to system crashes. The maximum velocity (MAX Vel) and maximum acceleration (Max Acc) of the planned trajectory are recorded in Tab.~\ref{tab:Tracking Error of Agile Transportation}. Each trajectory is repeatedly executed $5$ times to calculate the root-mean-square error (RMSE) of trajectory tracking. The results indicate that our control scheme can successfully track the agile trajectory up to the limit of thrust that can be provided by the practical aerial robot in the MARTS. Besides, for the MARTS, its allowable agility decreases as the payload's mass increases.

\begin{table}[t]
    \centering
    \caption{Ablation Study of the Force Compensation}
    \label{tab:ablation study of force compensation}
    \renewcommand{\arraystretch}{1.25}
    \begin{tabular}{|c|cc|cc|}
    
    \hline
     & \multicolumn{2}{c|}{INDI}       & \multicolumn{2}{c|}{ Reference Force}    \\ \hline
    \makecell[c]{Max Acc \\ ($m/s^2$)} & \multicolumn{1}{c|}{\makecell[c]{RMSE \\($cm$) }}  & \makecell[c]{MAXE \\ ($cm$)} & \multicolumn{1}{c|}{\makecell[c]{RMSE \\ ($cm$)}} & \makecell[c]{MAXE\\($cm$)} \\ \hline
$4.9$     & \multicolumn{1}{c|}{$\bm{3.910}$}     &   $\bm{7.141}$   & \multicolumn{1}{c|}{{7.898}}     &  11.713   \\ \hline
$6.3$     & \multicolumn{1}{c|}{$\bm{4.129}$}     &   $\bm{7.449} $  & \multicolumn{1}{c|}{8.214}     &  14.667   \\ \hline
$7.7$     & \multicolumn{1}{c|}{$\bm{4.383}$}     &   $\bm{9.315}$  & \multicolumn{1}{c|}{8.435}     &  17.399   \\ \hline
$9.1$     & \multicolumn{1}{c|}{$\bm{4.674}$}     &   $\bm{10.267}$  & \multicolumn{1}{c|}{$\usym{2717}$}     &  $\usym{2717}$   \\ \hline
    \end{tabular}
    \end{table}

\begin{figure}[t]
    \begin{center}
        \includegraphics[width=0.95\columnwidth]{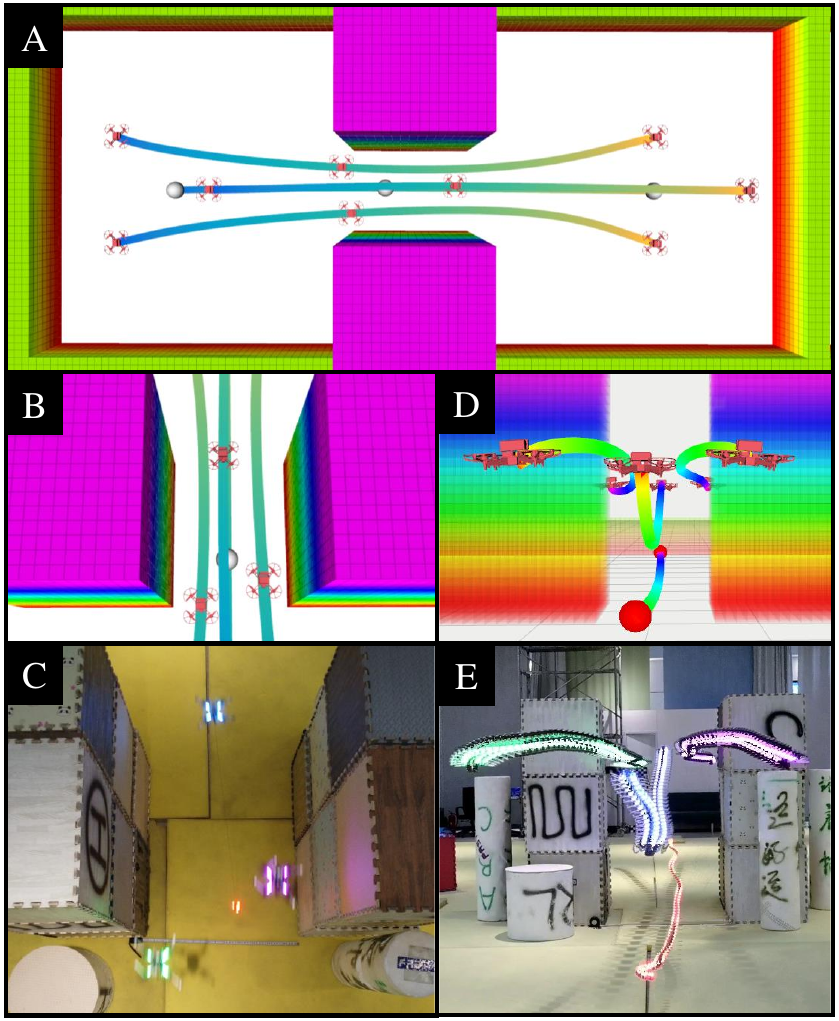}
    \end{center}
    \vspace{-0.3 cm}
    \caption{Illustration of the MARTS passing through a narrow gap. (A) A simulation shows the whole optimized trajectory for safely passing through the gap by regulating the relative position among the aerial robots. (B) A top-view in rviz at the moment of passing through the gap. (C) The snapshot of experiment w.r.t (B). (D) The side-view of the optimized trajectory in rviz. (E) The sequential snapshots of experiment w.r.t (D).}
    \label{fig:narrow gap}
\end{figure}

\subsubsection{Scenario 3. Comparative Study of Cable's Force Compensation}
In this experiment, we evaluate the effect of accurate force estimation and compensation in outer-loop control of each aerial robot on control error. We replace the accurate force compensation based on INDI in the outer-loop control of the proposed control scheme with the reference force provided by the trajectory as a comparative control scheme. Four trajectories with different maximum velocities are planned for this comparative study. Then we transport a weight weighing $100g$ along each trajectory $5$ times using both the control schemes for a comparison. For each trajectory, each control scheme's maximum error (MAXE) along the trajectory and the RMSE are given in Tab.~\ref{tab:ablation study of force compensation}. We find that both the RMSE and MAXE of the control scheme using reference force are larger than the proposed control scheme. The control scheme using reference force cannot even successfully execute the trajectory with a maximum acceleration of $9.1m/s^2$. For the control scheme using reference force, the inevitable tracking error of payload induces the actual cable's force to periodically deviate from the reference force, leading to oscillation in the positions of aerial robots. This oscillation in turn further worsens the tracking precision of the payload. However, the proposed control scheme can effectively avoid this oscillation since it accurately compensates the actual cable's force.    
Therefore, this experiment validates the importance of the accurate force compensation based on INDI to substitute the reference force from the trajectory in the proposed control scheme for agile transportation.

\subsubsection{Scenario 4. Robustness Against Payload's Uncertainties}
Since there always exists an uncertainty in the estimation of the payload's mass and an uncertainty induced by the unavoidable swing around the cables' attaching point,  we test the robustness of the proposed control scheme against these uncertainties. In this experiment, we use a weight, an express carton, and a water bottle respectively as the payload, all of which weigh $200g$. First, we use $200g$ as the payload's mass to plan a trajectory, which is denoted by $\pm0\%$. Then, we increase and decrease $200g$ by $10\%, 20\%$, and $30\%$ as the payload's masses for the trajectory planning to simulate the imprecise estimations for the payload's mass and plan six trajectories. We adjust the optimization parameters to ensure the maximum velocities of all these $7$ trajectories are $3.5m/s$. Each payload tracks each trajectory $5$ times to calculate the RMSE. For all these tests, the cable cannot be attached to the payload's CoM strictly and all the payloads are not mass points, which implies that there exists an unavoidable swing of the payload around the attaching point. The RMSEs of these tests are recorded in Tab.~\ref{tab:Tracking Error of Payload with Uncertainties}. From the first column, the RMSEs of the water bottle and the carton are greater than the RMSE of the weight, which shows the detrimental effect of the unavoidable swing induced by the non-mass point payload on the control precision. Besides, for all three payloads, the RMSEs increase with the estimation errors of the payload's mass. Nevertheless, all the payloads can be successfully transported by our MARTS using all these trajectories planned by the imprecise payload's masses, which validate the robustness of the proposed control scheme against the uncertainties on the payload. 

\begin{table}[t]
    \caption{RMSEs of Multiple Payloads with Uncertainties}
    \label{tab:Tracking Error of Payload with Uncertainties}
    \centering
    
    \renewcommand{\arraystretch}{1.25}
    \begin{tabular}{|c|c|c|c|c|}
    \hline
     & $\pm 0\%$   & \begin{tabular}[c]{@{}c@{}} $+10\%$\\ $-10\%$ \end{tabular} & \begin{tabular}[c]{@{}c@{}}$+20\%$\\ $-20\%$\end{tabular} & \begin{tabular}[c]{@{}c@{}}$+30\%$\\ $-30\%$\end{tabular} \\ \hline
      Weight        & 4.472 & \begin{tabular}[c]{@{}c@{}}4.563\\ 4.804\end{tabular} & \begin{tabular}[c]{@{}c@{}}4.747\\ 4.867\end{tabular} & \begin{tabular}[c]{@{}c@{}}4.897 ($\uparrow 9.5\%$) \\ 5.158($\uparrow 15.3\%$)\end{tabular} \\ \hline

    Carton        & 4.935($\uparrow 10.4\%$) & \begin{tabular}[c]{@{}c@{}}5.243\\ 5.054\end{tabular} & \begin{tabular}[c]{@{}c@{}}5.424\\ 5.142\end{tabular} & \begin{tabular}[c]{@{}c@{}}5.493($\uparrow 22.8\%$) \\ 5.450($\uparrow 21.9\%$)\end{tabular} \\ \hline

     Bottle & 5.039($\uparrow 12.7\%$) & \begin{tabular}[c]{@{}c@{}}5.227\\ 5.132\end{tabular} & \begin{tabular}[c]{@{}c@{}}5.386\\ 5.383\end{tabular} & \begin{tabular}[c]{@{}c@{}}5.575($\uparrow 24.7\%$)\\ 5.499($\uparrow 23.0\%$) \end{tabular} \\ \hline
    \end{tabular}
    \end{table}

% \usepackage{multirow}

% \usepackage{multirow}

% \usepackage{multirow}

\begin{figure*}[t]
    \begin{center}
        \includegraphics[width=1.88\columnwidth]{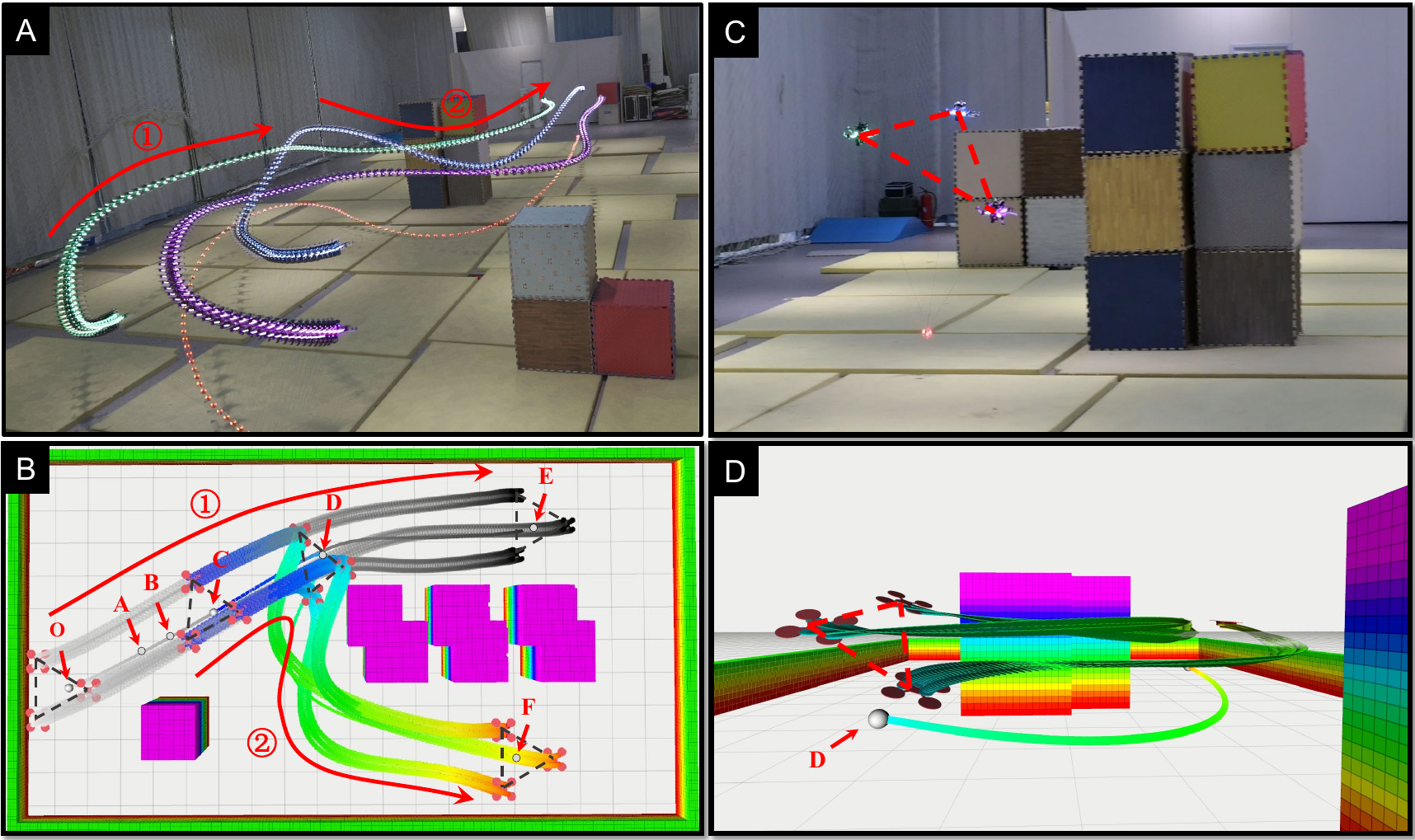}
    \end{center}
    \vspace{-0.3 cm}
    \caption{Illustration of an emergent replanning for the MARTS towards a new target. (A) Actual experiment and the trajectories of the MARTS. (B) The simulation schematic for this emergency replanning. (C) Special zoomed-in view of the MARTS for the frame at point D. (D) The side-view of the simulation schematic for this emergency replanning.}
    \label{fig:emergent replanning}
\end{figure*}

\begin{figure*}[t]
    \begin{center}
        \includegraphics[width=1.88\columnwidth]{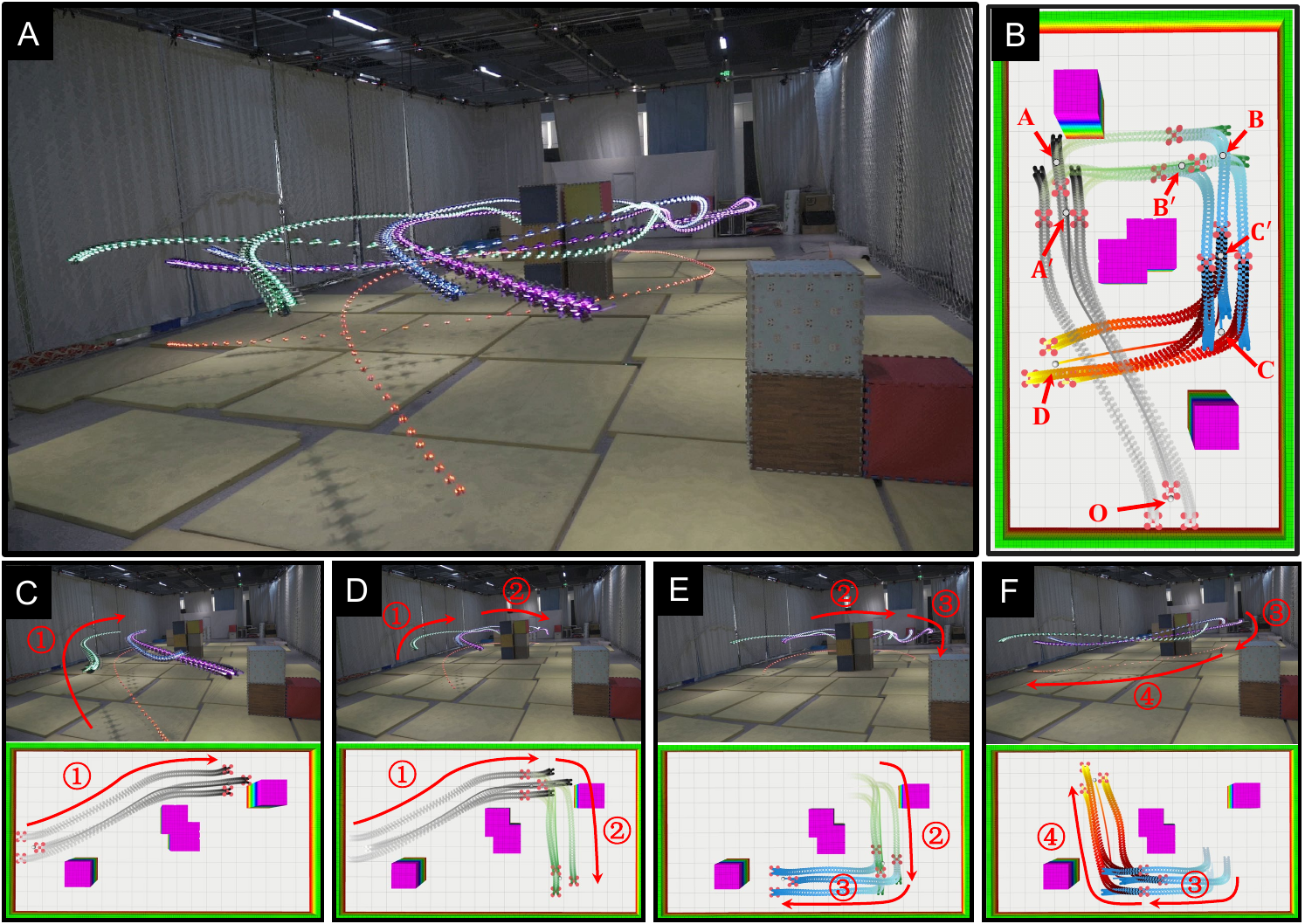}
    \end{center}
    \vspace{-0.3 cm}
    \caption{Illustration of consecutively replanning for the MARTS towards new targets. (A) Sequential snapshot of the MARTS in the experiment. (B) Simulation schematic for consecutively replanning. (C) The planned trajectory $\widetilde{OA}$ for the first target point $A$. (D) The replanned trajectory $\widetilde{A'B}$ for the second target point $B$. (E) The replanned trajectory $\widetilde{B'C}$ for the third target point $C$. (F) The replanned trajectory $\widetilde{C'D}$ for the fourth target point $D$.}
    \label{fig:consecutively replanning}
\end{figure*}

\subsection{System Scheme Validation in Various Scenarios}
We further construct a variety of scenarios to validate the performance of our MARTS, including the safety of agile transportation in complex environments, the agility of trajectory in narrow spaces, and the responsiveness of trajectory planning. The details are described as follows. 

\subsubsection{Continuous S-shaped Turns}
An experimental environment with multiple consecutive S-shaped turns is constructed as shown in Fig.~\ref{fig:head}B. In this experiment, a safe and agile trajectory of $24.43m$ is planned just using $162ms$. The MARTS transports a payload of $200g$ smoothly through these consecutive S-shaped turns and reaches the target point, without colliding with any obstacle.   This experiment validates the capability of our planning scheme to generate a safe and agile trajectory in real time.

\subsubsection{Narrow Gap}
We construct a narrow gap with a width of $1m$ smaller than the initial size ($1.6m$) of the MARTS. In this experiment, our planning scheme contracts the lateral size of the MARTS by optimizing all the cables' directions to ensure safe transportation through the narrow gap. As shown in Fig.~\ref{fig:narrow gap}, the result validates that our planning scheme can generate an agile trajectory by adjusting the configuration of MARTS to ensure the system's safety, which enhances its adaptability to narrow spaces. This experiment verifies the agility of the trajectory planned by our planning scheme.

\subsubsection{Emergent Replanning Towards a New Target}
% In autonomous navigation scenarios without a priori map,
% since the MARS needs to continually update the target point from the front-end safe path for generating a new local trajecotry at the back-end. 
In many missions, the responsiveness of replanning is important for mission efficiency. To validate the responsiveness of our planning scheme, we artificially construct an emergent replanning scenario as shown in Fig.~\ref{fig:emergent replanning}A. This test consists of the following five phases. 1. The trajectory $ \widetilde{OE}$ is generated after the point $E$ is selected as the target. 2. The MARTS flies along the trajectory $\widetilde{OE}$, and we select a new target point $F$ for the MARTS when the payload reaches point $A$. 3. The planning module takes the state at point $C$, which is $100ms$ ahead of point $A$, as the initial state and replans the trajectory. 4. The MARTS continually flies along the trajectory $\widetilde{AE}$ until it receive a new trajectory $\widetilde{CF}$ at the point $B$. 5. At point $C$, the MARTS switches the trajectory from trajectory $\widetilde{OE}$ to $\widetilde{CF}$. Thanks to the real-time capability, our planning module just uses $72ms$ to replan a new trajectory, which depresses the elapsed time required for replanning and thus strives for a larger safe distance away from the obstacle. Besides, an agile trajectory is generated by our planning scheme so that the MARTS can bypass the obstacle using a large maneuver and successfully switch the trajectory topology within this limited safe distance, which further improves the efficiency of MARTS. Fig.~\ref{fig:emergent replanning}C-D are the snapshots of this maneuver.

\subsubsection{Consecutively Replanning}
Finally, to further validate the responsiveness of our planning scheme for replanning, we serially select the target $A$, $B$, $C$, and $D$ around the obstacle for the MARTS to construct a consecutively replanning scenario, as shown in Fig.~\ref{fig:consecutively replanning}. All the replannings are carried out within $50ms\sim80ms$ and continuous state transitions are guaranteed for smooth trajectory switching at points $A'$, $B'$, and $C'$. The result of this experiment, as well as the result of the previous experiment sufficiently verify the responsiveness of our planning scheme. For more details, please watch the experimental video.

\section{Conclusion}
\label{sec:conclusion}
In this paper, we carefully derive the flatness maps for the aerial robot in the MARTS subject to dynamics coupling with payload and kinematic constraints provided by cable. A real-time trajectory planning scheme is proposed to generate safe, dynamically feasible, and agile trajectories for the MARTS in complex environments. A robust and distributed control scheme is proposed to track agile trajectory without relying on the closed-loop control and state measurement for both the payload and cable, even if there exists uncertainty on the payload. Finally, we deploy a practical MARTS containing three aerial robots with adequate simulations and experiments to validate the effectiveness of our planner and control schemes and exhibit great application value.

\bibliography{main}

\end{document}